\documentclass[runningheads]{llncs}

\usepackage[mobile]{eccv}

\usepackage{eccvabbrv}

\usepackage{graphicx}
\usepackage{booktabs}

\usepackage[accsupp]{axessibility}  

\usepackage[dvipsnames]{xcolor}

\newcommand{\xmark}{\ding{55}}

\usepackage{amsmath}
\usepackage{amssymb}
\usepackage{multirow}
\usepackage{adjustbox}
\usepackage{rotating}
\usepackage{comment}
\usepackage{pifont}
\usepackage{boldline}

\usepackage[pagebackref,breaklinks,colorlinks,citecolor=eccvblue]{hyperref}

\usepackage{orcidlink}
\newcommand{\greencheckmark}{\textcolor{green}{\checkmark}}
\newcommand{\redxmark}{\textcolor{red}{\xmark}}

\begin{document}

\title{Pseudo-keypoint RKHS Learning for Self-supervised 6DoF Pose Estimation} 

\titlerunning{RKHSPose}

\author{Yangzheng Wu\orcidlink{0000-0001-8893-0672}  \and 
Michael Greenspan\orcidlink{0000-0001-6054-8770}
}

\authorrunning{Y.Wu and M. Greenspan}

\institute{RCVLab, Dept. of Electrical and Computer Engineering, Ingenuity Labs, \\ Queen's University, Kingston, Ontario, Canada \\\email{ \{y.wu, michael.greenspan\}@queensu.ca}}

\maketitle
\begin{abstract}
We address the simulation-to-real domain gap in 
six degree-of-freedom pose estimation (6DoF PE), 
and propose a novel self-supervised keypoint voting-based 6DoF PE framework, effectively narrowing this gap using a learnable kernel in RKHS.
We formulate this domain gap as a distance in high-dimensional feature space,
distinct from previous iterative matching methods.
We propose an adapter network,
which is pre-trained on purely synthetic data with synthetic ground truth poses,
and which evolves the network parameters
from this source synthetic domain to the target real domain.
Importantly, the real data training only uses pseudo-poses estimated by pseudo-keypoints, and thereby requires no real ground truth data annotations.
Our proposed method is called
RKHSPose,
and achieves state-of-the-art performance 
among self-supervised methods
on three commonly used 6DoF PE datasets including LINEMOD ($+4.2\%$), Occlusion LINEMOD ($+2\%$), and YCB-Video ($+3\%$).
It also compares favorably to fully supervised methods 
on all six applicable BOP core datasets, 
achieving within $-11.3\%$
to $+0.2\%$ of the top fully supervised results.


\end{abstract}    
\section{Introduction}
\label{sec: Introduction}

RGB-D Six Degree-of-Freedom Pose Estimation (6DoF PE) 
is a problem being actively explored in computer vision research.
Given an RGB image 
and its associated depth map, 
the task is to detect scene objects 
and estimate their poses comprising 
3DoF rotational angles and  3DoF translational offsets
in the camera reference frame.
This task enables many applications such as augmented reality~\cite{hinterstoisser2012model,posecnn,hodan2018bop,kaskman2019homebreweddb}, robotic bin picking~\cite{doumanoglou2016recovering,hodan2017tless,kleeberger2019large}, 
autonomous driving~\cite{ma2019accurate,xiao2019identity} and image-guided surgeries~\cite{gadwe2018real,greene2023dvpose}.


\begin{figure}[t]
\begin{center}
\includegraphics[width=.95\columnwidth]{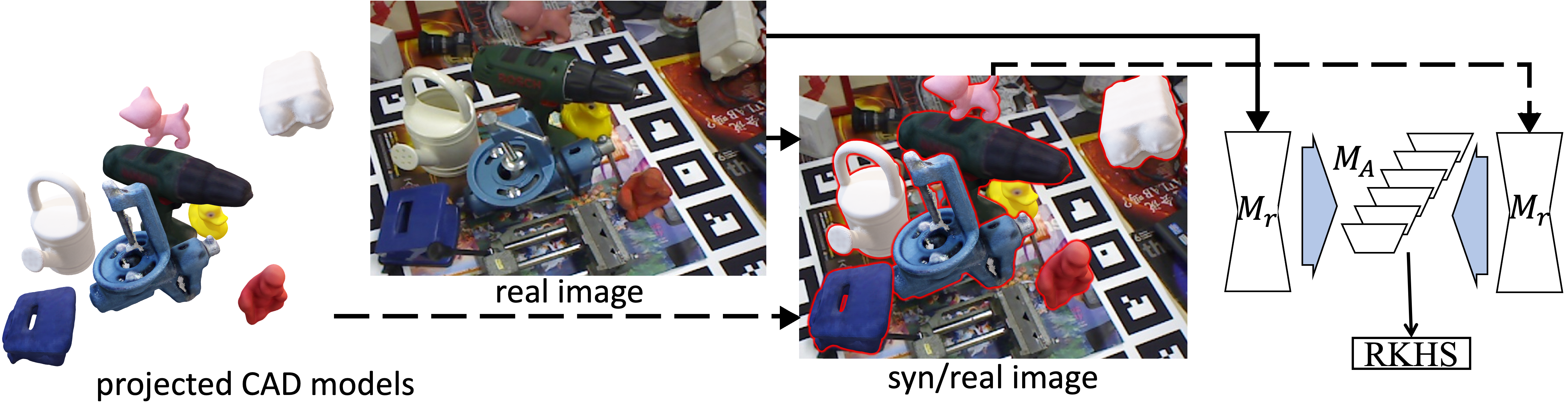}
\caption{
RKHSPose adapts the network pretrained on synthetic data 
to real test scenes (left), by comparing network feature spaces with real image inputs (solid arrows), against those with syn/real image (right) inputs (dashed arrows). $M_r$ regresses radial quantities, $M_A$ is the Adapter network, and RKHS maps features into a higher dimensional space. 
\label{fig:teaser}}
  
\end{center}
\end{figure}
As with other machine learning (ML) tasks,
fully supervised
6DoF PE
requires large annotated datasets.
This requirement is particularly 
challenging
for 6DoF PE, as the annotations
comprise not only the identity of the objects
in the scene,
but also their 6DoF pose, which makes the data
relatively expensive to annotate
compared to related tasks such as
classification, detection and segmentation.
This is due to the fact that humans are not able to qualitatively or intuitively estimate 6DoF pose, which therefore requires additional instrumentation of the scene at data collection, and/or more sophisticated user annotation tools~\cite{hinterstoisser2012model, posecnn}.
Consequently,
synthetic 6DoF PE datasets~\cite{hodan2018bop,kleeberger2019large} have been introduced,
either as an additional complement to real datasets, or as standalone purely synthetic datasets.
Annotated synthetic datasets are of course trivially inexpensive to collect,
simply because
precise synthetic pose annotation
in a simulated environment is fully automatic.
A known challenge in using synthetic data, however, is that there typically exists a domain gap between the real and synthetic data, which makes results less accurate when 
inferring in real data using models trained on
purely synthetic datasets. 
Expectations of the potential benefit of synthetic datasets has led to the exploration of a rich set of Domain Adaptation (DA) methods, which specifically aim to reduce the domain gap~\cite{bozorgtabar2020exprada, zhang2017curriculum,pinheiro2019domain}
using inexpensive synthetic data
for a wide variety of tasks,
recently including 6DoF PE~\cite{guo2023knowledge,lee2022uda,li2021sd}.


Early methods~\cite{wu2022vote,kleeberger2020single} ignored the simulation-to-real (sim2real) domain gap and
nevertheless improved performance by
training on both synthetic and real annotated data,
effectively augmenting the real images with the synthetic.
However, these methods still required real labels to be sufficiently accurate and robust for practical applications,
partially due to the domain gap.
As shown in Fig.~\ref{fig:teaser},
the rendered synthetic objects (right image) have a slightly different appearance than the real objects (left image).
The details of the CAD models, both geometry and texture, are not precise, as can be seen for the \emph{can} object (which lacks a mouth and shadows), the \emph{benchvise} (which is missing a handle), and the \emph{holepuncher} (which has coarse geometric resolution).
Several recent methods~\cite{kleeberger2021investigations,sundermeyer2018implicit,tan2023smoc,chen2023texpose,wang2020self6d} have
started to 
address the sim2real gap 
for 6DoF PE
by first training on labeled synthetic data 
and then fine-tuning on unlabeled real data.
Commonly known as \emph{self-supervised}, these methods
reduce the domain gap
by adding extra supervision
using features extracted from real images without requiring real ground truth (GT) labels.
The majority of these methods are viewpoint/template matching-based, and
the self-supervision commonly 
iteratively matches 2D masks or 3D point clouds~\cite{kleeberger2021investigations,sundermeyer2018implicit,wang2020self6d}.

While the above-mentioned self-supervison works have shown promise,
there exist a wealth of DA techniques that can be brought to bear to improve performance further for this task.
One such technique is Reproducing Kernel Hilbert Space (\emph{RKHS}), which
is a kernel method that has been shown to be effective for DA~\cite{khosravi2023existence,al2021learning,bietti2019kernel}.
RKHS was initially used to create decision boundaries for non-separable data~\cite{cortes1995support,pearson1901liii},
and has
been shown to be effective at reducing the domain gap for various tasks and applications~\cite{zhang2018aligning,shan2023unsupervised,chen2020homm}.
The reproducing kernel guarantees that the domain gap can be statistically measured, allowing 
network parameters trained on synthetic data to be effectively adapted to the real data, using specifically tailored metrics.


To address the sim2real domain gap in 6DoF PE, 
we propose RKHSPose,
which is a keypoint-based method~\cite{wu2022vote,wu2022keypoint} trained on a mixture of a large collection of labeled synthetic data, and a small handful of unlabeled real data.
RKHSPose estimates
the intermediate radial voting quantity,
which has been shown to be effective for estimating keypoints~\cite{wu2022vote},
by first training a modified FCN-Resnet-18
on purely synthetic data, with 
automatically labeled synthetic GT poses.
The radial quantity is a map of the distance from each image pixel to each keypoint.
Next, real images are passed through the synthetically trained network,
resulting in a set of pseudo-keypoints.
The real images and their corresponding pseudo-keypoints are used to render a
set of `synthetic-over-real' (\emph{syn/real}) images, by first estimating the pseudo-pose from the pseudo-keypoints, and then overlaying the synthetic object,
rendered with the pseudo-pose,
onto the real image. The network training
then continues on the syn/real images, invoking an RKHS network module with a trainable linear product kernel, which minimizes the Maximum Mean Discrepancy (MMD) loss. 
At the front end, a proposed 
keypoint radial voting network 
learns to cast votes to estimate keypoints
from the backend-generated radial maps.
The final pose is then determined using ePnP~\cite{lepetit2009ep} 
based on the estimated and corresponding object keypoints.


The main contributions of this work are:
\begin{itemize}
    \item A novel learnable RKHS-based Adapter backend network architecture to minimize the sim2real domain gap in 6DoF PE;
    \item A novel CNN-based frontend network 
    for keypoint radial voting;
    \item A self-supervised keypoint-based 6DoF PE method, RKHSPose, which is shown to have state-of-the-art (SOTA) performance, based on our experiments and several ablation studies. 
\end{itemize}

\section{Related Work}
\label{sec: Related Works}
\subsection{6DoF PE}

\begin{table}[t]
\begin{center}
\caption{Existing self-supervised 6DoF PE methods. Some methods use CAD models, labeled synthetic (syn) data (images+poses), and real data without GT labels, while others use only labeled synthetic data. Many methods require a ROI such as a bounding box or a semantic mask, manually labeled or estimated by an existing framework. SO-Pose~\cite{Di_2021_ICCV} used very few real labels to improve performance.
\label{tab:self supervised 6DoF PE}}
\begin{adjustbox}{max width=.9\columnwidth}
\begin{tabular}{lcccccc}
\toprule
\vspace{-3 pt}
\multirow{2}{*}{\vspace{0 pt} method} & \multirow{2}{*}{mode} & CAD & syn & real & real &\multirow{2}{*}{ROI} \\
 & & model & data & images & poses &  \\ \hline
SO-Pose~\cite{Di_2021_ICCV}&RGB&\greencheckmark & \greencheckmark & \greencheckmark & \greencheckmark&\greencheckmark\\ 
TexPose~\cite{chen2023texpose}&RGB&\greencheckmark & \greencheckmark & \greencheckmark & \redxmark&\greencheckmark\\ 
SMOC-Net~\cite{tan2023smoc}&RGB&\greencheckmark&\greencheckmark&\greencheckmark&\redxmark&\greencheckmark\\
FS6D~\cite{he2022fs6d} &RGBD &\greencheckmark & \greencheckmark & \redxmark  &\redxmark  &  \greencheckmark \\  
AAE~\cite{sundermeyer2018implicit}&RGB &\greencheckmark & \greencheckmark & \redxmark &\redxmark&\greencheckmark\\ 
MHP~\cite{Manhardt_2019_ICCV}&RGB &\greencheckmark & \greencheckmark & \redxmark & \redxmark&\greencheckmark \\ 
Sock et al.~\cite{sock2020introducing}&RGB &\greencheckmark & \greencheckmark & \redxmark & \redxmark&\greencheckmark\\ 
DSC~\cite{xiao2019pose}&RGBD &\greencheckmark & \greencheckmark & \redxmark & \redxmark&\greencheckmark \\ 
Sundermeyer~\cite{sundermeyer2020multi}&RGB &\greencheckmark & \greencheckmark & \redxmark  & \redxmark &\greencheckmark \\  
LatentFusion~\cite{park2020latentfusion}&RGB &\redxmark & \greencheckmark & \redxmark &\redxmark&\greencheckmark   \\
OSOP~\cite{shugurov2022osop}&RGBD &\greencheckmark & \greencheckmark & \redxmark & \redxmark&\redxmark\\ 
Kleeberger et al.~\cite{kleeberger2021investigations}&D&\greencheckmark & \greencheckmark & \redxmark & \redxmark&\redxmark\\ 
Su et al.~\cite{su2015render}&RGB&\greencheckmark & \greencheckmark & \greencheckmark & \redxmark&\redxmark\\ 
Self6D~\cite{wang2020self6d}&RGBD&\greencheckmark & \greencheckmark & \greencheckmark & \redxmark&\redxmark\\ 
Self6D++~\cite{wang2021occlusion}&RGB&\greencheckmark & \greencheckmark & \greencheckmark & \redxmark&\redxmark\\ 
Deng et al.~\cite{deng2020self}&RGBD&\greencheckmark & \greencheckmark & \greencheckmark & \redxmark&\redxmark\\ 
RKHSPose (Ours)&RGBD&\greencheckmark & \greencheckmark & \greencheckmark & \redxmark&\redxmark\\ 
\bottomrule

\end{tabular}
\end{adjustbox}

\end{center}
\end{table}

ML-based 6DoF PE methods~\cite{posecnn,oberweger2018making,pvnet} 
all train a network to 
regress quantities,
such as 
keypoints and camera viewpoints,
as have been used in classical algorithms~\cite{hinterstoisser2012model}.
ML-based methods,
which
initially became popular for the general object detection task,
have started to dominate the 6DoF PE literature
due to their accuracy and efficiency.

There are two main categories of ML-based fully supervised methods:
\emph{feature matching}-based~\cite{posecnn,oberweger2018making,su2022zebrapose,haugaard2022surfemb,hai2023rigidity}, and 
\emph{keypoint}-based methods~\cite{pvnet,pvn3d,he2021ffb6d,yang2023object}.
Feature matching-based methods
make use of the structures from
general object detection networks
most directly.
The network encodes and matches features and estimates pose
by either regressing elements of the pose
(e.g. the transformation matrix~\cite{tan2023smoc}, rotational angles and translational offsets~\cite{posecnn} or 3D vertices~\cite{su2022zebrapose}) directly,
or 
by regressing some intermediate feature-matching representations, such as viewpoints or segments.

In contrast, 
keypoint-based methods encode features
to estimate keypoints 
which are predefined
within the reference frame of an object's CAD model.
These methods then use (modified) classical algorithms
such as PnP~\cite{p3p,lepetit2009ep}, Horn's method~\cite{horn1988closed}, 
and ICP~\cite{besl1992icp}
to estimate the final pose
from corresponding image and model keypoints. 
Unlike feature-matching methods,
keypoint-based methods are typically more accurate due to
redundancies encountered 
through voting schemes~\cite{pvnet,wu2022vote,pvn3d} 
and by generating confidence hypotheses of keypoints~\cite{he2021ffb6d,yang2023object}.
Recently, self-supervised 6DoF PE methods have been explored in order to reduce the reliance on labeled real data,
which is expensive to acquire.
As summarized in Table~\ref{tab:self supervised 6DoF PE}, these methods commonly use real images without GT labels.
Some methods use pure synthetic data and CAD models only, with the exception of LatentFusion~\cite{park2020latentfusion} which trained the model using only synthetic data.
The majority of these methods~\cite{he2022fs6d,shugurov2022osop,sundermeyer2020multi,su2015render,wang2020self6d,deng2020self,wang2021occlusion,chen2023texpose,tan2023smoc,sock2020introducing, lin2022category}
are inspired by fully supervised feature-matching methods, except DPODv2~\cite{shugurov2021dpodv2} and DSC~\cite{xiao2019pose}, in which keypoint correspondences are
rendered and matched.

A few methods~\cite{shugurov2022osop,deng2020self}
fine-tuned the pose trivially by iteratively matching the template/viewpoint, whereas
others~\cite{xiao2019pose,kleeberger2021investigations,park2020latentfusion,he2022fs6d} augmented the training data by adding noise~\cite{xiao2019pose,kleeberger2021investigations}, rendering textures~\cite{he2022fs6d,shugurov2021dpodv2} and
creating a latent space~\cite{park2020latentfusion}.
Some methods~\cite{sundermeyer2018implicit,sundermeyer2020multi,Manhardt_2019_ICCV} also implemented DA techniques such as codebook encoding~\cite{sundermeyer2020multi}, Principle Component Analysis (PCA)~\cite{sundermeyer2020multi} and symmetric Bingham distributions~\cite{Manhardt_2019_ICCV}.
Most methods~\cite{shugurov2022osop,su2015render,wang2020self6d,wang2021occlusion,sock2020introducing,xiao2019pose} used rendering techniques to render and match a template.
There are a few methods that combined  3D reconstruction techniques, such as Neural radiance fields (Nerf)~\cite{mildenhall2021nerf} and Structure from Motion (SfM)~\cite{ullman1979interpretation}.
TexPose~\cite{chen2023texpose} matched CAD models to segments generated by Nerf,
and SMOC-Net~\cite{tan2023smoc} used SfM to create the 3D segment and matched with the CAD model.

\subsection{Kernel Methods and Deep Learning}
While deep learning is the most common ML technique within the computer vision literature,
kernel methods~\cite{cortes1995support,pearson1901liii,sejdinovic2013equivalence} have also been actively  explored.
Kernel methods are typically in RKHS space~\cite{szafraniec2000reproducing} with reproducing properties that facilitate solving
non-linear problems by
mapping input data into high dimensional spaces that can be linearly separated~\cite{ghojogh2021reproducing,corcoran2020end}.
Well-known early kernel methods that have been applied to computer vision are Support Vector Machines~\cite{cortes1995support} and PCA~\cite{pearson1901liii}. 
A recent method~\cite{sejdinovic2013equivalence} linked energy distance and MMD in RKHS, 
and showed the effectiveness of kernels in statistical hypothesis testing. 

More recent studies compare kernel methods with deep learning networks~\cite{khosravi2023existence,al2021learning,bietti2019kernel,bietti2019group}.
RKHS is found to 
 perform better on classification tasks 
than one single block of a CNN comprising convolution, pooling and downsampling (linear) layers~\cite{khosravi2023existence}. 
RKHS can also help with CNN generalization by meta-regularization on image registration problems\cite{al2021learning}.
Similarly, norms (magnitude of trainable weights) defined in RKHS help with  CNN regularization~\cite{bietti2019kernel,bietti2019group}.
Further, discriminant information in label distributions in unsupervised DA is addressed and RKHS-based Conditional Kernel Bures metric is proposed~\cite{luo2021conditional}.
Lastly, the connection between Neural Tangent Kernel and MMD is established,
and an efficient MMD for two-sample tests is developed~\cite{cheng2021neural}.

Inspired by the previous work,
RKHSPose applies concepts of kernel learning to keypoint-based 6DoF PE,
to provide an effective means to self-supervise a synthetically trained network on unlabeled real data.

\section{Method}
\label{sec:Method}
\begin{figure*}[t]

\begin{center}
\includegraphics[width=0.93\textwidth]{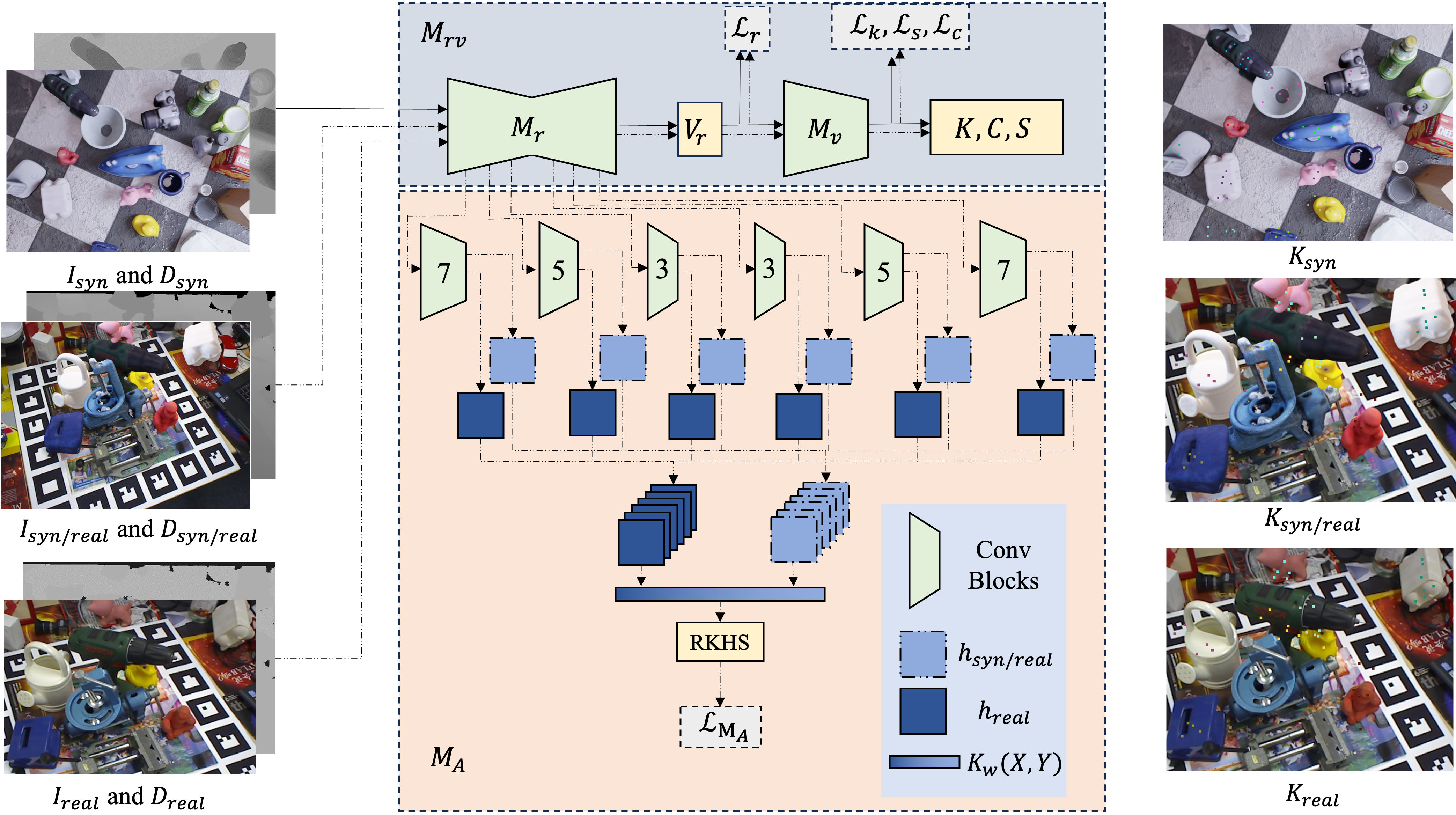}
\caption{
RKHSPose architecture. RKHPose is first trained on synthetic labeled data (solid arrows), and then finetuned on alternating syn/real and (unlabeled) real images (dashed arrows). $M_A$ is measured by MMD in RKHS by densely mapping the intermediate features of $M_r$ into high dimensional spaces with conv blocks. The distance is treated as $\mathcal{L}_{M_A}$ and back-propagated through $M_A$ and $M_r$.\label{fig:diagram}}
  
\end{center}
\end{figure*}
\subsection{Network Overview}

As shown in Fig.~\ref{fig:diagram}, RKHSPose is made up of two networks,
a main network $M_{rv}$ for keypoint regression and classification, 
and an Adapter network $M_A$. 
The input of $M_{rv}$ 
with shape $W\!\times\!H\!\times\!4$, 
is the concatenation of an RGB image $I$ and its corresponding depth map $D$.
This input can be synthetically generated with an arbitrary background ($I_{syn}$ and $D_{syn}$),
a synthetic mask overlayed on a real background ($I_{syn/real}$ and $D_{syn/real}$),
or a real ($I_{real}$ and $D_{real}$) image.
The outputs are $n$ projected 2D keypoints $K$, 
along with corresponding classification labels $C$ and confidence scores $S$. $K$ are organized into instance sets based on $C$ and geometric constraints.

$M_{rv}$ comprises two sub-networks, 
regression network $M_r$ and 
voting network $M_v$.
Inspired by recent voting techniques~\cite{pvnet,pvn3d,wu2022vote,wu2022keypoint,zhou2023deep},
$M_r$ estimates an intermediate voting quantity, 
which is a radial distance map $V_r$~\cite{wu2022vote}, 
by using a modified Fully Connected ResNet 18 (FCN-ResNet-18).
The radial voting map $V_r$, with shape $W\!\times\!H$,
stores the Euclidean distance
from each object point to each keypoint
in the 3D camera world reference frame.
The voting network $M_v$ 
(described in Sec.~\ref{sec: Keypoints ConvRVoting})
then takes $V_r$ as input, 
accumulates votes, and detects peaks to
estimate $K$, $C$ and $S$.

The Adapter network $M_A$ consists of
a series of CNNs 
which encode pairs of feature maps from $M_{rv}$,
and are trained on both synthetic overlayed and pure real data.
$M_A$ encodes
feature map pairs $(f_{syn/real}, f_{real})$ 
into corresponding high-dimensional feature maps $(h_{syn/real}, h_{real})$.
The input data, 
$(I_{syn/real},I_{real})$ and $(D_{syn/real},D_{real})$,
are also treated as $(f_{syn/real}, f_{real})$ during the learning of $M_A$.
Each of these networks creates a high-dimensional latent space,
essentially the Reproducing Kernel Hilbert Space (RKHS)~\cite{aronszajn1950theory}
and contributes to the learning of both $M_{rv}$ and $M_A$ 
by calculating the 
MMD~\cite{gretton2012kernel}.
While RKHS has been applied effectively to other DA tasks, to our knowledge, this is the first time that it has been applied to 6DoF PE, and the adapter network architecture is novel.

The loss function $\mathcal{L}$ is made up of five elements:
Radial regression loss $\mathcal{L}_{r}$ for $V_r$;
Keypoint projection loss $\mathcal{L}_{k}$ for $K$;
Classification loss $\mathcal{L}_{c}$ for $C$; 
Confidence loss $\mathcal{L}_{s}$ for $S$, and finally;
Adapter loss $\mathcal{L}_{A}$ for the comparison of intermediate feature maps.
The regression losses $\mathcal{L}_{r}$, $\mathcal{L}_{k}$, and $\mathcal{L}_{s}$ all use the smooth L1 metric, whereas classification loss $\mathcal{L}_{c}$ uses the cross-entropy metric $H(\cdot)$.
RKHSPose losses can then be denoted as:
\begin{eqnarray}
\mathcal{L}_{r} 
&=& 
smooth_{\mathcal{L}_1}
(V_r, 
\widehat{V}_r
)
\label{eq:Lr}
\\ 
\mathcal{L}_{k} &=& smooth_{\mathcal{L}_1}(K, \widehat{K})
\label{eq:Lk} \\
\mathcal{L}_{c} &=& H(p(C), 
\widehat{p}(C))
\label{eq:Lc} \\
\mathcal{L}_{s} &=& smooth_{\mathcal{L}_1}(S, \widehat{S})
\label{eq:Ls} \\
\mathcal{L}_{A} &=& MMD(
\widehat{f}_{syn/real}, 
\widehat{f}_{real}
)
\label{eq:LMA} \\
\mathcal{L} = \lambda_r \mathcal{L}_{r}
&+&
\lambda_k \mathcal{L}_{k} + \lambda_c \mathcal{L}_{c} + \lambda_s \mathcal{L}_{s} + \lambda_A \mathcal{L}_{A}
\label{eq:L}
\end{eqnarray}
\noindent where
$\lambda_r$, $\lambda_k$, $\lambda_c$, $\lambda_s$, and $\lambda_A$ are weights for adjustment during training,
and all non-hatted quantities are GT values. 
At inference, $M_{rv}$ takes $I_{real}$ and $D_{real}$ as input, and
 outputs  $K$, $C$ and $S$.
The keypoints $K$ are ranked and grouped by $S$ and $C$, and are then forwarded into the ePnP~\cite{lepetit2009ep} algorithm which estimates 6DoF pose values. 
ICP can then be optionally applied using the depth data to refine the estimated pose.

\subsection{Convolutional RKHS Adapter}
\label{sec: Convolutional RKHS Discriminator}
Reproducing Kernel Hilbert Space $\mathcal{H}$ is a commonly used vector space for Domain Adaptation~\cite{pan2010domain,venkateswara2017deep}.
Hilbert Space is a complete metric space 
(in which every Cauchy sequence of points has a limit within the metric) 
represented by the inner product of vectors.
For a non-empty set of data $\mathcal{X}$,
a function $\mathcal{K}_{\mathcal{X}}\!:\!\mathcal{X}\!\times\!\mathcal{X}\!\rightarrow\!\mathcal{R}$ is a reproducing kernel if:
%
\begin{equation}
\begin{cases}
    &k(\cdot,x) \in \mathcal{H}\;\;\; \forall \;\;\; x \in \mathcal{X} \\
    & \left\langle f(\cdot), k(\cdot,x)\right\rangle = f(x) \;\;\; \forall \;\;\; x \in \mathcal{X}, f \in \mathcal{H}
\end{cases}
\label{eq:cases}
\end{equation}
where $\left\langle a, b \right\rangle$
denotes the inner product of two
vectors $a$ and $b$,
$k(\cdot,x)=\mathcal{K}_\mathcal{X}$ 
for each $x\!\in\!\mathcal{X}$, 
and $f$ is a function in $\mathcal{H}$.
The second equation in
Eq.~\ref{eq:cases}
is an expression of the reproducing property of $\mathcal{H}$.

In order to utilize $\mathcal{H}$ for DA,
we expand the kernel definition to 
two sets of data $\mathcal{X}$ and $\mathcal{Y}$.
The reproducing kernel
 can then be defined as:
\begin{equation}
    \mathcal{K}(\mathcal{X}, \mathcal{Y})=\left\langle \mathcal{K}_\mathcal{X}, \mathcal{K}_\mathcal{Y}\right\rangle_H
\label{eq:rkhs_k}
\end{equation}
Here, 
$\mathcal{K}_\mathcal{X}$ and $\mathcal{K}_\mathcal{Y}$ are 
themselves inner product kernels
that map $\mathcal{X}$ and $\mathcal{Y}$ respectively into their own Hilbert spaces,
and 
$ \mathcal{K}(\mathcal{X}, \mathcal{Y})$
is the joint Hilbert space kernel
of $\mathcal{X}$ and $\mathcal{Y}$.
Note that $ \mathcal{K}(\mathcal{X}, \mathcal{Y})$ is not the kernel commonly defined for CNNs. 
Rather, it is the similarity function defined in kernel methods, such as is used by Support Vector Machine (SVM) techniques to calculate similarity measurements.
Some recent methods~\cite{misiakiewicz2022learning,mairal2014convolutional,mairal2016end,chen2020convolutional} named it a Convolutional Kernel Network (CKN) to distinguish it from CNNs.

Various kernels of CKN, such as 
the Gaussian Kernel~\cite{mairal2014convolutional}, the RBF Kernel~\cite{mairal2016end} and the Inner Product Kernel~\cite{misiakiewicz2022learning}, 
have been shown to be comparable to shallow CNNs for various tasks, especially for Domain Adaption.
The most intuitive inner product kernel 
is mathematically similar to 
a fully connected layer, 
where trainable weights are multiplied by the input feature map.
To allow the application of RKHS methods into our CNN based Adapter network $M_A$, 
trainable weights are added to $\mathcal{K}(\mathcal{X}, \mathcal{Y})$~\cite{liu2020learning}.
The trainable Kernel $\mathcal{K}_{w}(\mathcal{X}, \mathcal{Y})$ can then be denoted as:
\begin{equation}
    \mathcal{K}_{w}(\mathcal{X}, \mathcal{Y}) = \left \langle \left\langle \mathcal{X},W_\mathcal{X}\right\rangle,\left\langle \mathcal{Y},W_\mathcal{Y}\right\rangle \right\rangle_H
\label{eq:kw}
\end{equation}
where $W_\mathcal{X}$ and $W_\mathcal{Y}$ are trainable weights. 
By adding $W_\mathcal{X}$ and $W_\mathcal{Y}$ , $K_{w}(\mathcal{X}, \mathcal{Y})$
still satisfies the
RKHS constraints.

The sim2real domain gap of feature maps $f$ being trained in $M_{rv}$ (with few real images and no real GT labels)
are hard to measure using trivial distance metrics.
In contrast, RKHS can be a more accurate and robust space for comparison, since
it is known to be capable of handling high-dimensional data with a low number of samples~\cite{ghorbani2020neural,misiakiewicz2022learning}.
To compare $(f_{syn/real}, f_{real})$ in RKHS,
a series of CNN layers encodes $(f_{syn/real}, f_{real})$ into higher-dimensional features $(h_{syn/real}, h_{real})$,
followed by the trainable $\mathcal{K}_w(\mathcal{X}, \mathcal{Y})$.
Once mapped into RKHS, 
$h_{syn/real}\! = \!\{sr_i\}_{i=1}^{m}$
and 
$h_{real}\! = \!\{r_i\}_{i=1}^{m}$
can then be measured by Maximum Mean Discrepancy (MMD), a common DA measurement~\cite{gretton2012kernel,liu2020learning}, which is the square distance between the kernel embedding~\cite{gretton2006kernel}:
\begin{align}
& MMD(h_{syn/real}, h_{real}) =
\frac{1}{m}
\left[\left(\sum_{i=1}^{m}\sum_{j=1}^{m}{k_w(sr_i,sr_j)}\right.\right.
\!\!-\!\! 
\left.\sum_{i=1}^{m}{k_w(sr_i,sr_i)}\right)
\nonumber \\
&-\left(\sum_{i=1}^{m}\sum_{j=1}^{m}{k_w(sr_i,r_j)} \right.
\!\!-\!\! 
\left.\sum_{i=1}^{m}{k_w(sr_i,r_i)}\right)
+\left(\sum_{i=1}^{m}\sum_{j=1}^{m}{k_w(r_i,r_j)}\right.
\!\!-\!\! 
\left.\left.\sum_{i=1}^{m}{k_w(r_i,r_i)}\right)\right]^{\frac{1}{2}}
\label{eq:mmd2}
\end{align}
where $k_w()$ is the feature element of $\mathcal{K}_w()$.


In summary, the Adapter $M_A$ shown in Fig.~\ref{fig:diagram}
measures MMD 
for each $(h_{syn/real}, h_{real})$ in RKHS,
by increasing the $f_{syn/real}$ and $f_{real}$ dimension using a CNN and 
thereby constructing a learnable kernel $\mathcal{K}_w$.
The outputs are the 
feature maps
$h_{syn/real}$ and $h_{real}$, 
which are supervised by loss
 $\mathcal{L}_{A}$
during training of the real data epochs.
Based on the experiments in Sec.~\ref{sec: Discriminator Kernels Metrics},
our trainable inner product kernel is
shown to be more accurate
for our task
than other known kernels~\cite{mairal2014convolutional,mairal2016end,misiakiewicz2022learning} that we tested, that are often used in such kernel methods.

\subsection{Keypoint Radial Voting Network}
\label{sec: Keypoints ConvRVoting}
The network $M_v$ votes for keypoints using a CNN architecture, taking the radial voting quantity
$V_r$
resulting from $M_r$
as input.
VoteNet~\cite{qi2019deep} previously used a CNN approach to vote for object centers,
whereas other keypoint-based techniques have implemented
GPU-based
parallel RANSAC~\cite{pvn3d,pvnet} methods for offset and vector quantities.
The radial quantity is known to be more accurate 
than the vector or offset quantities, and 
has been previously implemented with a CPU-based
parallel 
accumulator space method~\cite{wu2022vote,wu2022keypoint}.
Given its superior accuracy, 
$M_v$ implements radial voting using a CNN to
improve efficiency.
Given a 2D radial map $\widehat{V}_r$ 
estimated by $M_r$ and supervised by GT radial maps $V_r$, 
the task is to accumulate votes, find the peak, and estimate the keypoint location.
The  
$V_r$ foreground pixels (which lie on the target object)
store the Euclidean distance from these pixels
to each of the keypoints, with background (non-object) pixels set to value -$1$. 

The estimated radial map $\widehat{V}_r$ is indeed an inverse heat map of the candidate keypoints' 
locations, distributed in a radial pattern centered at the keypoints. 
To forward $\widehat{V}_r$ into a CNN voting module, 
it is inversely normalized
so that it becomes a heat map. 
Let $v^{max}_r$ and $v^{min}_r$ be 
the maximum and minimum global radial distances for all objects in a dataset,
which can be calculated by iterating through all GT radial maps,
or alternately generated from the object CAD models.
An inverse radial map $\widehat{V}^{-1}_r$ can then be denoted as
$\widehat{V}^{-1}_r=(v^{max}_r\!-\!\widehat{V}_r)/(v^{max}_r\!-\!v^{min}_r)$.
Voting network $M_v$ takes $\widehat{V}^{-1}_r$ as the input and 
generates the accumulated vote map 
by a series of convolution, ReLu, and batch normalization layers.
The complete network architecture is provided in the Supplementary material Sec. S.2.
The background pixels are filtered out by a ReLu layer, 
and only foreground pixels contribute to voting.
The accumulated vote map is then max-pooled for peak extraction,
and reshaped using a fully connected layer into a $n\! \times\! 4$ output. 
The output represents $n$ keypoints
and comprises
$n\!\times\! 2$ projected 2D keypoints $K$, 
$n$ classification labels $C$, 
and $n$ confidence scores $S$.
The labels $C$ indicate which object the corresponding keypoint belongs to, and $S$ ranks the confidence level of keypoints before being forwarded into ePnP~\cite{lepetit2009ep}
for pose estimation.
$M_v$ is supervised by $\mathcal{L}_k$, $\mathcal{L}_c$, and $\mathcal{L}_s$ (Eqs.~\ref{eq:Lk}-\ref{eq:Ls})
and is trained end-to-end along with $M_r$.

\section{Experiments}
\subsection{Datasets and Evaluation Metrics}
\label{sec:Datasets and Evaluation Metrics}
RKHSPose uses BOP  Procedural Blender~\cite{Denninger2023} (PBR) synthetic images~\cite{hodan2018bop,hodavn2020bop,sundermeyer2023bop,denninger2020blenderproc} for the synthetic training phase.
The images are generated 
by dropping synthetic objects (CAD models) onto a plane 
in a simulated environment using PyBullet~\cite{coumans2021}, and then rendering them
with synthetic textures.
All objects in the synthetic images are thus automatically labeled with precise GT poses. 
We evaluated RKHSPose for the six BOP~\cite{hodavn2020bop,sundermeyer2023bop} core datasets (LMO~\cite{hinterstoisser2012model}, YCB~\cite{posecnn}, TLESS~\cite{hodan2017tless}, TUDL~\cite{hodan2018bop}, ITODO~\cite{drost2017introducing}, and HB~\cite{kaskman2019homebreweddb}), all except IC-BIN~\cite{doumanoglou2016recovering}, which does not include any real training or validation images and is therefore not applicable.
ITODO and HB have no real images in the training set, and so
for training 
we instead used the real images in their validation sets, which were disjoint from their test sets.

Our main results are evaluated with the ADD(S)~\cite{hinterstoisser2012model} 
metric for the LM and LMO dataset, 
and the ADD(S) AUC~\cite{posecnn} metric for the YCB dataset.
These are the standard metrics commonly used to compare 
self-supervised 6DoF PE methods.
ADD(S) is based on
the mean distance (minimum distance for symmetry) between the 
object surfaces for GT and estimated poses,
whereas ADD(S) AUC plots a curve formed by ADD(S)
for various object diameter thresholds.
We use the BOP average recall ($AR$) metrics for our ablation studies.
The $AR$ metric, based on the original ADD(S)~\cite{hinterstoisser2012model}, evaluates three aspects,
including Visible Surface Discrepancy ($AR_{VSD}$),
Maximum Symmetry-Aware Surface/Projection Distance ($AR_{MSSD}$ and $AR_{MSPD}$)~\cite{hodan2018bop}.

\subsection{Implementation Details}
RKHSPose is trained on a server with an Intel Xeon 5218 CPU and two RTX6000 GPUs with a batch size of 32.
The Adam optimizer is used for the training of $M_{rv}$, 
on both synthetic and real data,
and $M_A$ is optimized by SGD.
Both of the optimizers have an initial learning rate of $lr\!=\!1e$-$3$ and weight decay $1e$-$4$ for $80$ and $20$ epochs respectively.

The input of the network is normalized before training, as follows.
The RGB images $I$ are normalized and standardized using ImageNet~\cite{Imagenet} specifications.
The depth maps are each individually normalized by their local minima and maxima, to lie within a range of 0 to 1.
The radial distances in radial map $V_r$ and 2D projected keypoints are both normalized by the width and height of $I$.

A single set of four keypoints is chosen by KeyGNet~\cite{wu2023learning} for the set of all objects in each dataset.
One extra \emph{background} class is added to $C$ 
in order to filter out the redundant background points in $K$.
$M_{rv}$ is first trained for $120$ epochs on synthetic data, during which $M_A$ remains frozen. 
Following this, training proceeds for an additional $80$ epochs which alternate between real and synthetic data. When training on real data, both $M_A$ and $M_{rv}$ weights are learned, whereas $M_A$ is frozen for the alternating synthetic data training.

During real data epochs,
$M_{rv}$ initially 
estimates pseudo-keypoints for each real image. 
These pseudo-keypoints are then forwarded into ePnP for pseudo-pose estimation.
Each estimated pseudo-pose 
is then augmented into a set of poses $P_{aug}$
by applying arbitrary rotational 
and translational perturbations
with 
respective 
ranges of $[-\frac{\pi}{18}, \frac{\pi}{18}]$ radians
and 
$[-0.1 , 0.1]$ 
along three axes 
within the normalized model frame,
which is 
defined using the largest object in the dataset.
The set of syn/real images
are rendered
by overlaying onto the real image the CAD model of each object 
using each augmented pose value in
$P_{aug}$.
The cardinality of
$P_{aug}$
is set to be one less than the batch size, and
the adapter $M_A$
is trained on a mini-batch of 
the hybrid images resulting from $P_{aug}$, 
plus the image resulting from the original estimated pseudo-pose.


Initially
$M_A$ is frozen,
and
for the first $80$ epochs,
the loss is set to emphasize the classification and the visibility score regression, i.e. $\lambda_c\!=\!\lambda_s\!=\!0.6$ and $\lambda_r\!=\!\lambda_k\!=\!0.4$.
Following this, up to epoch $200$, the scales of losses are then exchanged to fine-tune the localization of the keypoints, i.e.
$\lambda_c\!=\!\lambda_s\!=\!0.4$ and $\lambda_r\!=\!\lambda_k\!=\!0.6$.
After epoch $120$,
$M_A$ is unfrozen each alternating epoch, and 
$\lambda_D$ is set to 1 during the remaining $M_A$ training epochs.
This training strategy, shown in Fig.~\ref{fig:diagram},
minimizes the sim2real gap
without 
any real image GT labels, and using very few (320) real images.  

\begin{table}[t]
\vspace{-1\baselineskip}
\begin{center}
\caption{Comparison with other methods. 
Accuracy of RKHSPose for LM and LMO is evaluated with ADD(S), and for
YCB is evaluated with ADD(S) AUC. 
All  `Supervision: Syn + Self' methods use real images without real labels. 
\label{tab:mainresults}
 }
\begin{adjustbox}{max width=\columnwidth}
\begin{tabular}{lcccccc}
\cline{4-7}
\multicolumn{3}{c}{}
&
\multicolumn{4}{c}{Dataset/Metric} 
\\
\toprule
\multicolumn{1}{c}{\multirow{3}{*}{Method}} &\multicolumn{2}{c}{\multirow{2}{*}{Real data}} & LM  & LMO  & \multicolumn{2}{c}{YCB}\\
\multicolumn{1}{c}{}&\multicolumn{2}{c}{} &\multicolumn{1}{c}{}&&ADD(S)&ADD-S\\
  &\multirow{-2}{*}{image}&\multirow{-2}{*}{label}&\multicolumn{2}{c}{\multirow{-2}{*}{ADD(S)}}  & AUC & AUC\\\hline 
  \multicolumn{7}{l}{
  \textbf{
  \hspace{.25 cm} Supervision: Syn (lower bound)}}\\
AAE &\redxmark&\redxmark&31.4&-&-&-\\
MHP &\redxmark&\redxmark&38.8&-&-&-\\
GDR (TexPose version)
&\redxmark&\redxmark&77.4&52.9&-&-\\
Self6D++ &\redxmark&\redxmark&77.4&52.9&77.8&89.4\\
Self6D++ with $D_{ref}$&\redxmark&\redxmark&88.0&62.5&79.2&90.1\\
Ours&\redxmark&\redxmark&78.2&54.3&76.5&90.2\\
Ours+ICP&\redxmark&\redxmark&87.9&55.7&78.3&91.3\\\hline
\multicolumn{7}{l}{
\textbf{
\hspace{.25 cm} Supervision: Syn + Self}}\\
Sock \emph{et al.} &\greencheckmark&\redxmark&60.6&22.8&-&-\\
DSC &\greencheckmark&\redxmark&58.6&24.8&-&-\\
Self6D &\greencheckmark&\redxmark&58.9&32.1&-&-\\
SMOC-Net &\greencheckmark&\redxmark&91.3&63.3&-&-\\
Self6D++ &\greencheckmark&\redxmark&88.5&64.7&80.0&91.4\\
TexPose &\greencheckmark&\redxmark&91.7&66.7&-&-\\
Ours &\greencheckmark&\redxmark&\underline{95.8}&\underline{68.6}&\underline{82.8}&\underline{92.4}\\
Ours+ICP &\greencheckmark&\redxmark&\textbf{95.9}&\textbf{68.7}&\textbf{83.0}&\textbf{92.6}\\ \hline
\multicolumn{7}{l}{
\textbf{
\hspace{.25 cm} Supervision: Syn + Real GT (upper bound)}}\\
SO-Pose&\greencheckmark&\greencheckmark&96.0&62.3&83.9&90.9\\
Self6D++&\greencheckmark&\greencheckmark&91.0&74.4&82.6&90.7\\
Ours &\greencheckmark&\greencheckmark&\underline{96.7}&\underline{70.8}&\underline{85.4}&\underline{92.2}\\
Ours+ICP &\greencheckmark&\greencheckmark&\textbf{96.8}&\textbf{71.3}&\textbf{85.6}&\textbf{92.4}\\ 
\bottomrule
\end{tabular}
\end{adjustbox}
\end{center}
\end{table}

\begin{table}[t]
\begin{center}
\caption{Comparison with fully supervised methods.
RKHSPose results on TLESS ($-1.8$), TUDL ($-0.4$), ITODD ($-4.6$) and HB ($+0.1$) compares to SOTA methods with full supervision of real GT labels.
Methods annotated with $^*$ use the detection results from other detection methods.
\label{tab:bop_main}
}
\begin{adjustbox}{max width=\columnwidth}
\begin{tabular}{lcccccccc}
\toprule
\multicolumn{1}{c}{\multirow{2}{*}{Method}} 
& real &
\multicolumn{7}{c}{Dataset} \\
&label&LM&LMO&TLESS&TUDL&ITODD&HB&YCB
  \\\hline 
  SurfEmb$^*$~\cite{haugaard2022surfemb}&\greencheckmark&-&76.0&82.8&85.4&65.9&86.6&79.9\\
  RCVPose3D~\cite{wu2022keypoint}&\greencheckmark&-&72.9&70.8&\textbf{96.6}&\textbf{73.3}&86.3&84.3\\ 
  RADet~\cite{yang2023object}+PFA$^*$~\cite{hu2022perspective}&\greencheckmark&-&\textbf{79.7}&85.0&96.0&67.6&86.9&\underline{88.8}\\
  ZebraPose~\cite{su2022zebrapose}&\greencheckmark&-&\underline{78.0}&\textbf{86.2}&95.6&65.4&92.1&\textbf{89.9}\\
Ours&\redxmark&95.7&68.2&85.5&\underline{96.2}&68.6&\underline{92.2}&83.6
\\
Ours+ICP&\redxmark&95.8&68.4&\underline{85.6}&\underline{96.2}&\underline{68.7}&\textbf{92.3}&83.8
\\
\bottomrule

\end{tabular}
\end{adjustbox}

\end{center}

\end{table}

\begin{figure*}[t]

\begin{center}
\includegraphics[width=\textwidth]{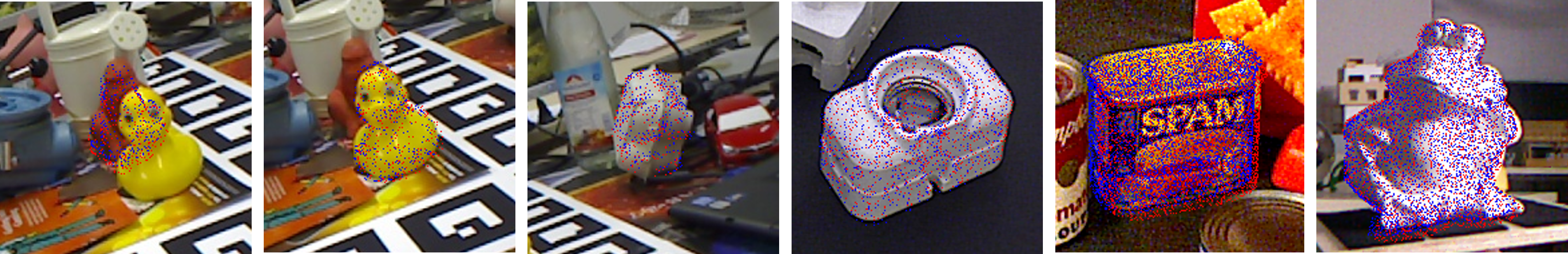}
\caption{
Qualitative overlay results on selected images. Red dots and blue dots are projected surface points from GT poses and estimated poses, respectively. 
\label{fig:results}}
  
\end{center}
\end{figure*}

\subsection{Results}
\label{sec: Results}

The results are summarized in Tables~\ref{tab:mainresults} and~\ref{tab:bop_main}.
To our knowledge, 
RKHSPose outperforms all existing self-supverised 6DoF PE methods.
In Table~\ref{tab:mainresults},
the upper bound fully supervised, lower bound synthetically supervised,
 and middle self-supervised methods (including ours) are compared.
On LM and LMO, our ADD(S) is $+4.2\%$ and $+2\%$ better than the second best method TexPose~\cite{chen2023texpose}.  
We compared our performance to Self6D++~\cite{wang2021occlusion},
which was the only other method that evaluated using YCB,
and saw a $3\%$ improvement after ICP on ADD(S) AUC.
Last but not least, 
in Table~\ref{tab:bop_main} our performance evaluated on the other four BOP core datasets 
is comparable to several upper bound methods 
from the BOP leaderboard.
Some test scenes with RKHSPose results are shown in Fig.~\ref{fig:results}.

RKHSPose runs at 34 fps on 
an Intel i7 2.5GHz CPU and an RTX 3090 GPU with 24G VRAM.
It takes on average $8.7$~ms for loading data, $4.5$~ms for forward inference through $M_{rv}$ ($\times10$ faster compared to the analytical radial voting in RCVPose~\cite{wu2022vote}), and $16.2$~ms for ePnP.

\section{Ablation Studies}
\label{sec: Ablation Studies}

\subsection{Dense Vs. Sparse Adapter}
\label{sec: Dense VS. Sparse Discriminator}
The Adapter $M_A$ densely matches the intermediate feature maps, whereas the majority of other methods~\cite{tan2023smoc,deng2020self,wang2020self6d} only compare the final output.
To show the benefits of dense comparison,
we conduct an experiment with different
variations of $M_A$.
A sparse matching $M^s_A$ network is trained on synthetic and real data
comparing only 
a single feature map,
which is the intermediate radial map.
$M^s_A$ has the exact same overall learning capacity (number of 
parameters) as the dense matching $M_A$ described in Sec.~\ref{sec: Keypoints ConvRVoting}. The results in Table~\ref{tab:discriminator desntiy}
show that
$M_A$ surpassed $M^s_A$ on all six datasets tested.
Specifically, on ITODD, $M_A$ is $12.1\%$ more accurate than $M^s_A$.
This experiment shows the effectiveness of our densely matched $M_A$.
\subsection{Syn/Real Synchronized Training}
\label{sec: Syn-Real Synchronized Training}
When training RKHSPose, real epochs are alternated with synthetic epochs. 
In contrast, some other methods~\cite{chen2023texpose,wang2020self6d,wang2021occlusion,tan2023smoc} separate the synthetic/real training.
We conducted an experiment to compare these two different training strategies,
the results of which are shown in Table~\ref{tab:train strategy}.
The alternating training performs slightly better ($+2.5\%$ on average) than the sequential training, possibly 
due to the early access to real scenes thereby avoiding local minima.

\subsection{Adapter Kernels and Metrics}
\label{sec: Discriminator Kernels Metrics}
\begin{table}[t]
\small
\begin{minipage}[t]{.47\linewidth}
\centering
\caption{
$AR$ of different adapters on LM and five BOP core datasets.\label{tab:discriminator desntiy}}
\begin{adjustbox}{max width=\columnwidth}
\begin{tabular}{ccccc}
\toprule
Adapter  & $AR_{VSD}$ & $AR_{MSSD}$ & $AR_{MSPD}$ & $AR$\\ \hline
$M^s_A$ &78.1&77.8&77.8&77.9\\
$M_A$ & 84.9&84.1&84.3&\textbf{84.4}\\
\bottomrule
\end{tabular}
\end{adjustbox}
\end{minipage}
\hspace{0.05\linewidth}
\begin{minipage}[t]{.47\linewidth}
\centering
\caption{
$AR$ of different training strategies on LM and five BOP core datasets.\label{tab:train strategy}}
\begin{adjustbox}{max width=\columnwidth}
\begin{tabular}{ccccc}
\toprule
Training Type& $AR_{VSD}$ & $AR_{MSSD}$ & $AR_{MSPD}$ & $AR$\\ \hline
Mixed &84.9&84.1&84.3&\textbf{84.4}\\
Sequantial&82.3&81.7&81.7&81.9\\\bottomrule
\end{tabular}
\end{adjustbox}
\end{minipage}
\end{table}
\begin{table}[t]
\small
\begin{minipage}[t]{.48\linewidth}
\centering
\caption{
[R Tab A] $AR$ of different kernels on LM and five BOP core datasets.\label{tab:kernels}}
\begin{adjustbox}{max width=\columnwidth}
\begin{tabular}{cccccc}
\toprule
Kernel&$w$ & $AR_{VSD}$ & $AR_{MSSD}$ & $AR_{MSPD}$ & $AR$\\ \hline
\multirow{+2}{*}{Linear}& \redxmark& 71.6&	70.8&	70.6&	71.0\\
& \greencheckmark&84.9&84.1&84.3&\textbf{84.4}\\
\multirow{+2}{*}{RBF} & \redxmark &  73.4&	72.9&	73.2&	73.2 \\
  &\greencheckmark&82.5&81.3&81.5&81.6\\\bottomrule
\end{tabular}
\end{adjustbox}
\end{minipage}
\begin{minipage}[t]{.47\linewidth}
\centering
\caption{
$AR$ of different metrics on LM and five BOP core datasets.\label{tab:metrics}}
\begin{adjustbox}{max width=\columnwidth}
\begin{tabular}{ccccc}
\toprule
Metric & $AR_{VSD}$ & $AR_{MSSD}$ & $AR_{MSPD}$ & $AR$\\ \hline
MMD &84.9&84.1&84.3&\textbf{84.4}\\
KL Div&78.0&77.8&78.0&77.9\\
Wass&80.9&80.6&80.9&80.8\\ \bottomrule
\end{tabular}
\end{adjustbox}
\end{minipage}

\end{table}
We use a linear (dot product) kernel and MMD in RKHS for domain gap measurements.
There are various other kernels and similarity measurements that can be implemented in RKHS as
described in Sec.~\ref{sec: Convolutional RKHS Discriminator}. 
First, we add trainable weights to the radial basis function (RBF) kernel in a similar manner as $K_w$ defined in Eq.~\ref{eq:kw}. 
The trainable RBF kernel on two sets of data $X$ and $Y$ is denoted as:
\begin{equation}
    K_{rbf}(X, Y) = exp(-w\left \|  X-Y\right \|^2)
\end{equation}
where $w$ are the trainable weights,
which replaces the original adjustable parameter in the classical RBF kernel.
We also experiment with the classical RKHS kernel functions without trainable weights,
including the inner product kernel and the original RBF kernel~\cite{vert2004primer},
for comparison.
Further, we experiment on other commonly used distance measures,
including Kullback-Leibler Divergence (KL Div, i.e. relative-entropy) and Wasserstein (Wass) Distance.

In Table~\ref{tab:kernels}, the RBF kernel performs similar to the linear product kernel with a slight performance dip.
In Table~\ref{tab:metrics}, MMD minimizes the domain gap better than the Wass Distance, followed by the KL Div metric, leading to a better overall performance on $AR$.
Based on these results, we used the linear product kernel with trainable weights and selected MMD as the main loss metric.

\begin{figure}[t]
\begin{center}
\includegraphics[width=0.7\columnwidth]{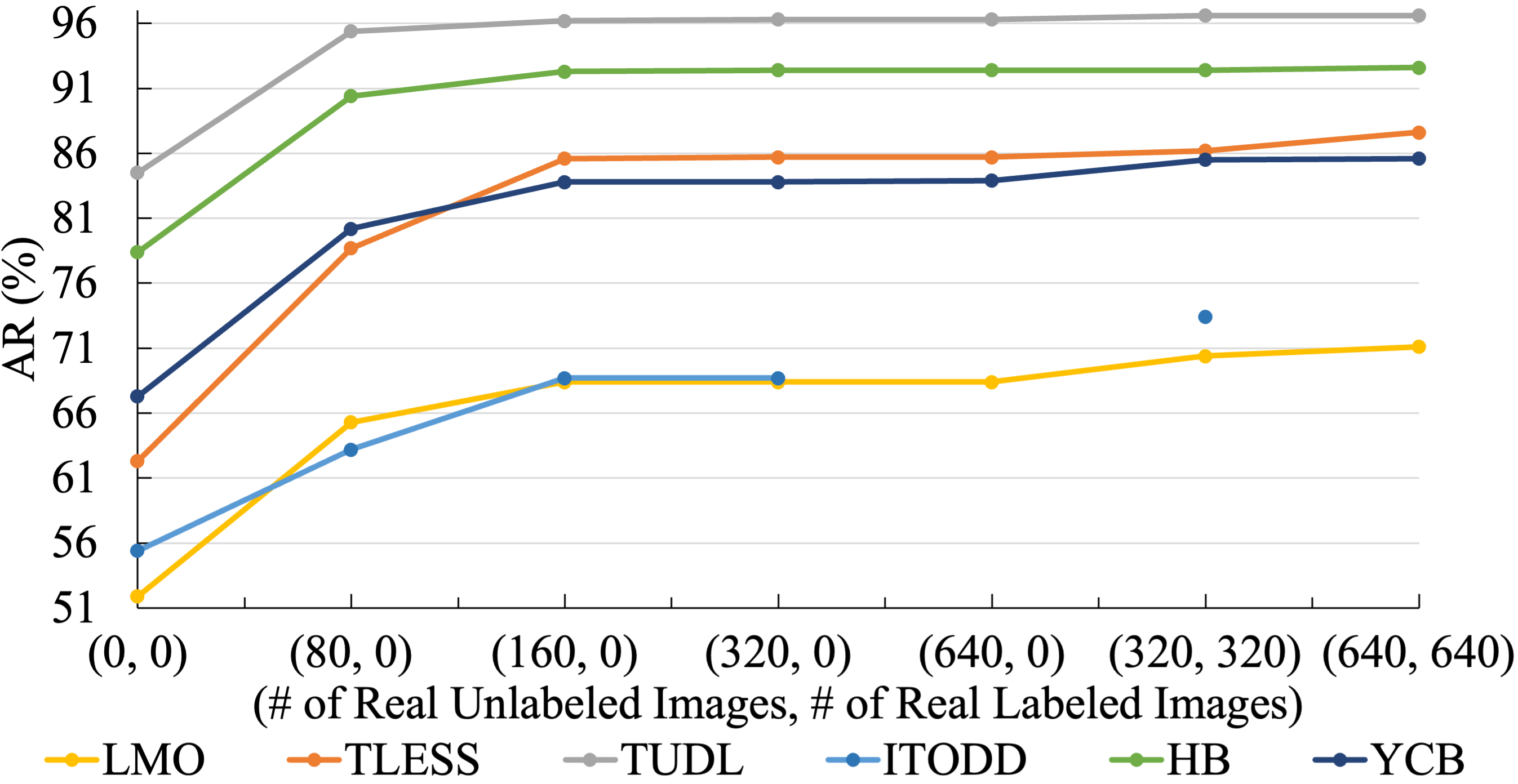}
\caption{
Impact of \# of real images with/without GT labels used during training. All datasets are evaluated by the BOP AR metric. 
We conduct experiments from 0 to 640 real images on all datasets,
except ITODD which contained only 357 real images.
\label{fig:noofreal}}
  
\end{center}

\end{figure}


\subsection{Number of Real Images and Real Labels}
\label{sec: No. of Real Images}

The objective of RKHSPose is to reduce real data usage and train without any real GT labels.
To show the effectiveness of the approach,
we conducted an experiment by training on different numbers of real images, 
the results of which are shown in Fig.~\ref{fig:noofreal}.
We used up to 640 real images in all cases, 
except for that of ITODD which contains
only 357 real images.
The $AR$ of all datasets saturates at 160 images except YCB. 
The further improvement of YCB beyond 160 images is also only $+0.1\%$ and saturates after $320$ real images. 
We nevertheless 
use $320$ real unlabeled images for our main results.
This experiment showed that adding more than 320 real labeled images did not significantly improve performance.

\section{Conclusion}
    
To sum up, we propose a novel self-supervised keypoint radial voting-based 6DoF PE method using RGB-D data called RKHSPose.
RKHSPose fine-tunes poses pre-trained on synthetic data by densely matching features with a learnable kernel in RKHS, 
using real data albeit without any real GT poses.
By applying this DA technique in feature space, 
RKHSPose achieved SOTA performance on the six applicable BOP core datasets, surpassing 
the performance of all other self-supervised methods.
Notably, the RKHSPose performance closely approaches that of several fully-supervised methods, which indicates the strength of the approach at reducing the sim2real domain gap for this problem.

\vspace{5mm}


\subsubsection*{Acknowledgements:}Thanks to Bluewrist Inc. and NSERC for their support
of this work.
\newpage
\let\theequationWithoutS\theequation
\renewcommand\theequation{S.\theequationWithoutS}
\let\thefigureWithoutS\thefigure 
\renewcommand\thefigure{S.\thefigureWithoutS}
\let\thetableWithoutS\thetable 
\renewcommand\thetable{S.\thetableWithoutS}
\let\thesectionWithoutS\thesection 
\renewcommand\thesection{S.\thesectionWithoutS}
\setcounter{page}{1}
\setcounter{section}{0}
\section{Supplementary Material Overview}
We document here some addition implementation details and results.
The detailed structures of keypoint radial voting network $M_v$ and Convolutional RKHS Adapter $M_A$ are described in Sec.~\ref{sec:Network DIagrams S} and shown in Figs.~\ref{fig:krv} and~\ref{fig:convrkhs}.
Sec.~\ref{sec:Radii Pattern for Keypoint Voting S} describes the visualized radial pattern (shown in Fig.~\ref{fig:v_rVSi_v_r}) of the dragon object in TUDL, which inspired us to use a CNN ($M_v$) to simulate the voting process.
The $AR_{VSD}$, $AR_{MSSD}$, $AR_{MSPD}$, and $AR$ results of all datasets we used for all ablation studies are summarized in Sec.~\ref{sec:Ablation Studies S} and listed in Fig.~\ref{fig:objdiam} and Tables~\ref{tab:discriminator desntiy supp},~\ref{tab:kernels supp},~\ref{tab:metrics supp}, and~\ref{tab:train strategy supp}.
The detailed ADD(S) results for each category of object on LM and LMO are listed in Tables~\ref{tab:LinemodFull} and~\ref{tab:OccLinemodFull}, and ADD-S AUC results on YCB are shown in Table~\ref{tab:YCBVideoFull}.
\section{Network Diagrams}
\label{sec:Network DIagrams S}
\begin{figure}[t]
     \centering
    \includegraphics[width=.9\columnwidth]{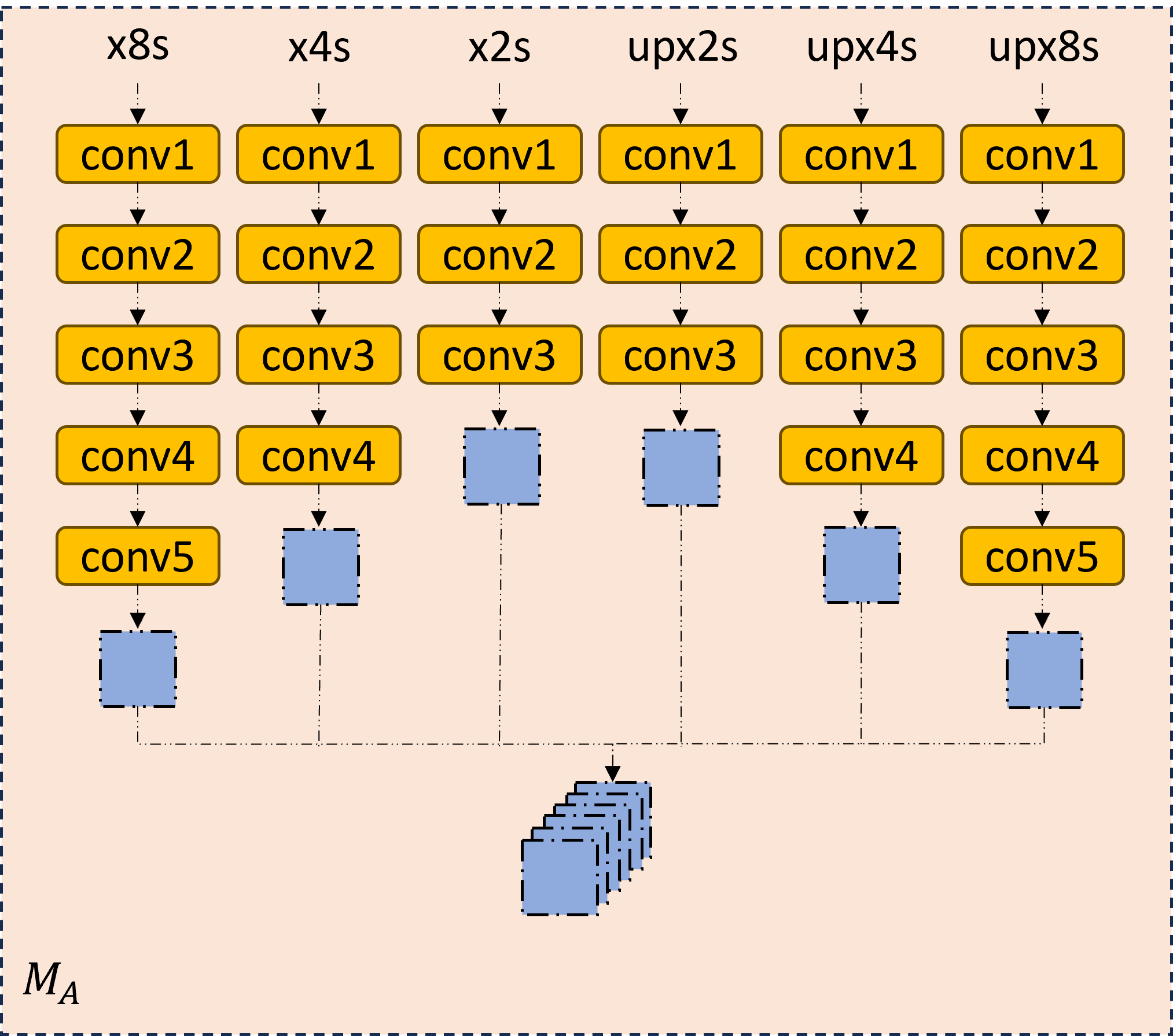}
     \caption{Convolutional RKHS Adapter $M_A$ detailed structure. 
     \label{fig:convrkhs}}
\end{figure}
\begin{figure}[h!]
     \centering
    \includegraphics[width=0.8\columnwidth]{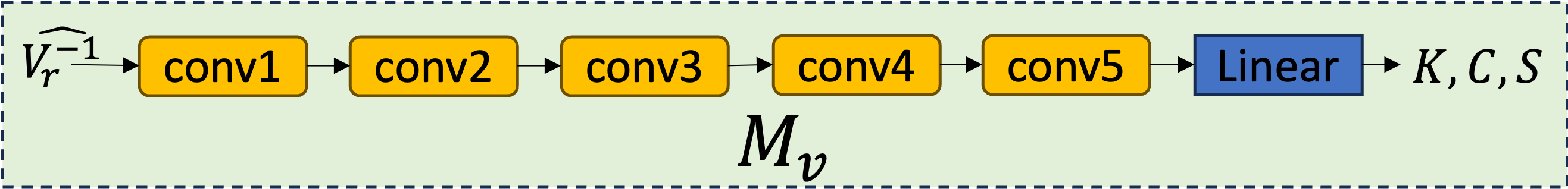}
     \caption{Keypoint radial voting network $M_v$ detailed structure. Each conv block comprises a $3\!\times\!3$ convolution, batch normalization, and ReLu layer. The final linear layer reshapes the feature map into the output $K, C, S$.\label{fig:krv}}
\end{figure}

The structure of Adapter network $M_A$
is shown in Fig.~\ref{fig:convrkhs}.
Each conv block comprises $3\!\times\!3$ convolution, batch normalization, and ReLu layers. Inputs of $M_A$, \ie x8s, x4s, x2s, upx2s, upx4s, and upx8s are feature maps extracted from each step of $M_r$. x8s stands for the 8th step of the encoding ResNet block, and upx8s is the 8th step of the decoding block, etc. Outputs of each column of conv blocks are concatenated and mapped into RKHS by a linear layer.
    
The keypoint radial voting network $M_v$ structure is shown in Fig.~\ref{fig:krv}.
The input of $M_v$ is the inversed radial map $\widehat{V}^{-1}_r$, defined in the main paper. $M_v$ comprises 5 convolution layers with a kernel size of $3$ and stride of $1$. A linear layer maps the feature map into the shape of $n\!\times\! 4$ comprising $n\!\times\! 2$ projected 2D keypoints $K$, 
$n$ classification labels $C$, 
and $n$ confidence scores $S$.

Estimated keypoints are organized into clusters based on geometric constraints when multiple instances of the same object appear within an image. More precisely, the estimated keypoints are grouped into instance sets by sorting the mean absolute differences of the Euclidean distances between keypoints defined on the CAD model, and those being estimated.

\clearpage
\pagebreak
\section{Radii Pattern for Keypoint Voting}
\label{sec:Radii Pattern for Keypoint Voting S}
\begin{figure}[h]
     \centering
        \begin{subfigure}[t]{0.45\textwidth}
         \centering
         \includegraphics[width=\textwidth]{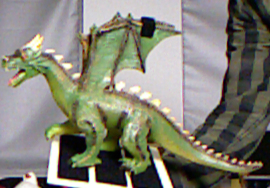}
         \caption{RGB image}
         \label{fig:v_r}
     \end{subfigure}
     \begin{subfigure}[t]{0.45\textwidth}
         \centering
         \includegraphics[width=\textwidth]{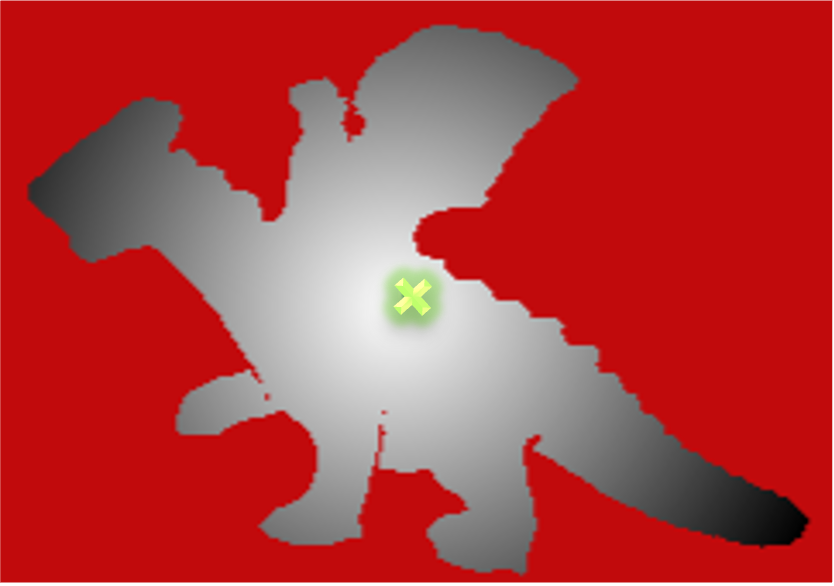}
         \caption{center keypoint $\widehat{V}^{-1}_r$}
         \label{fig:i_v_r_c}
     \end{subfigure}
        \begin{subfigure}[t]{0.45\textwidth}
         \centering
         \includegraphics[width=\textwidth]{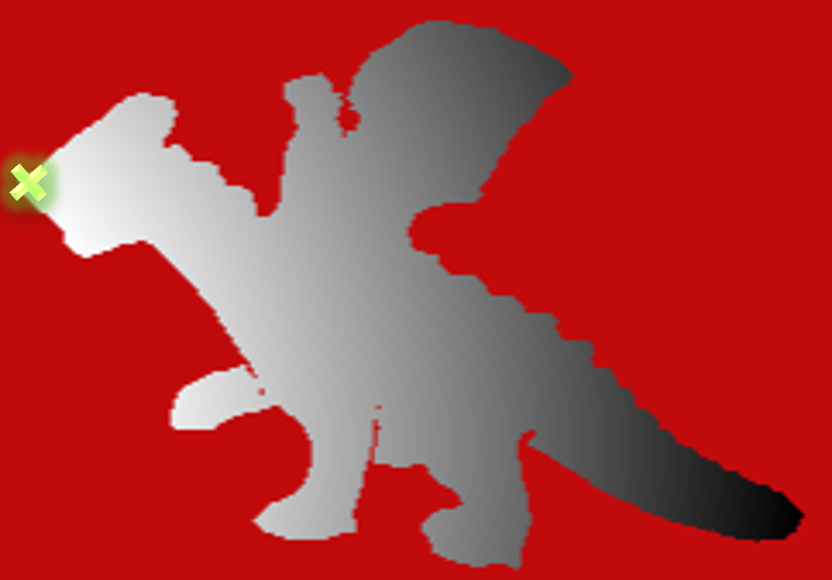}
         \caption{left keypoint $\widehat{V}^{-1}_r$}
         \label{fig:i_v_r_l}
     \end{subfigure}
     \begin{subfigure}[t]{0.45\textwidth}
         \centering
         \includegraphics[width=\textwidth]{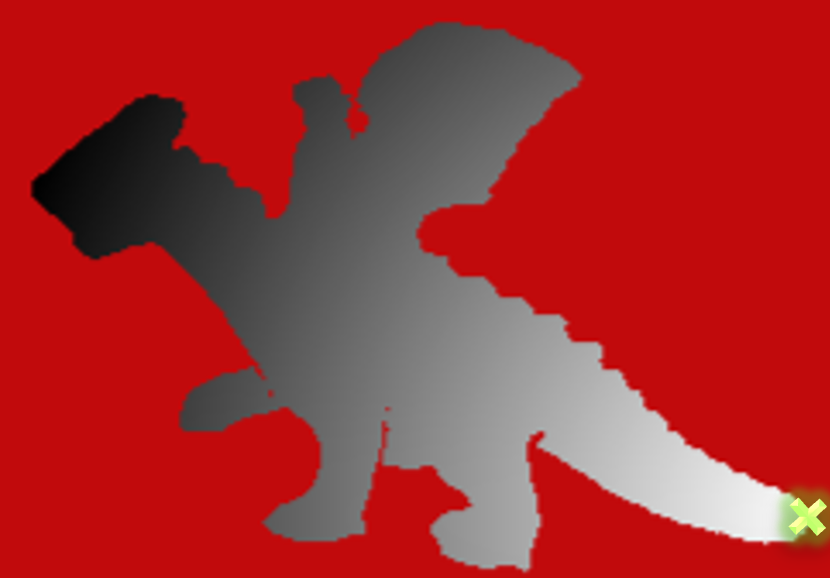}
         \caption{right keypoint $\widehat{V}^{-1}_r$}
         \label{fig:i_v_r_r}
     \end{subfigure}
     \caption{TUDL~\cite{hodan2018bop} dataset dragon object (a) RGB image and (b)-(d) inverse radial maps $\widehat{V}^{-1}_r$. The segmentation mask is applied to filter out the background, shown in red. Keypoints (pink stars) are located at the (b) center, (c) left, and (d) right. \label{fig:v_rVSi_v_r}}
\end{figure}

As shown in Fig.~\ref{fig:v_rVSi_v_r}, the estimated radial map $\widehat{V}_r$ is an inverse heat map of the candidate keypoints' locations 
, distributed in a radial pattern centered at the keypoints. 
The further away from the pixel to the keypoint,
$\widehat{V}_r$ has a  greater value.
Pixels with value $-1$
do not lie on an object.
This inspired us to use a CNN ($M_v$) to detect the peak, thereby localizing the keypoint.

\section{Ablation Studies}
\label{sec:Ablation Studies S}

The $AR_{VSD}$, $AR_{MSSD}$, $AR_{MSPD}$, and $AR$ results for all datasets used for all ablation studies are provided here. 

Table~\ref{tab:discriminator desntiy supp} shows the dataset-wise performance when sparse $M^{S}_{A}$ and dense $M_{A}$ RKHS Adapters are used to reduce the sim2real domain gap. $M_{A}$ performs better on all datasets than $M^{S}_{A}$.

Table~\ref{tab:kernels supp} demonstrates the impact of different kernels used in $M_{A}$ including linear and RBF kernels,
both with and without trainable weights.
$M_{A}$ with trainable linear kernel performs the best compared to others.

Table~\ref{tab:metrics supp} compares different RKHS metrics including MMD, KL Div, and Wass Distances.
$M_{A}$ evaluated by MMD outperforms the other metrics.

Table~\ref{tab:train strategy supp} illustrates the training strategies of the regression and voting model $M_{rv}$ and the adapter $M_{A}$, whether sequentially or mixed.
The mixed training is shown to lead to better (+2.5\%) performance.

\begin{figure}[b]
\begin{center}
\includegraphics[width=\columnwidth]{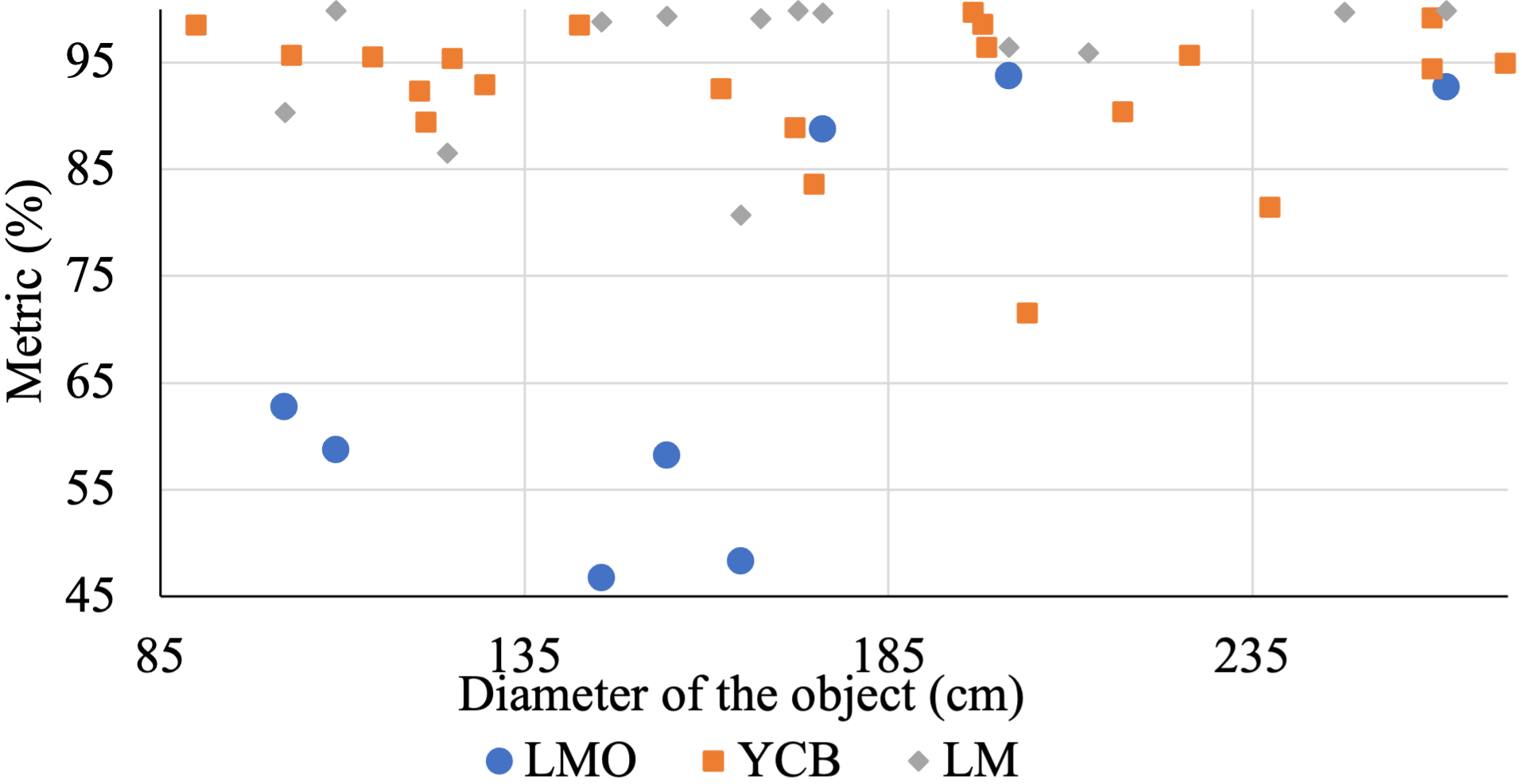}
\caption{Impact of object size (diameter) on accuracy, for three datasets (LM, LMO, and YCB). There is no noticeable performance change for different object sizes, except for small heavily occluded objects in the LMO dataset.
\label{fig:objdiam}}
  
\end{center}
\end{figure}

Lastly,
the correlation between object size and accuracy is also evaluated on LM, LMO, and YCB datasets, and the results are shown in Fig.~\ref{fig:objdiam}. 
The object size impact on the accuracy has no significant impact except when there are extreme occlusions in LMO.

\begin{table}[h]
\begin{center}
\caption{The impact of sparse ($M^s_A$) and dense ($M_A$) Adapters, defined in Sec. 5.1 in the main paper, on LM and LMO, and five BOP core datasets.\label{tab:discriminator desntiy supp}}
\begin{adjustbox}{max width=\columnwidth}
\begin{tabular}{cccccV{1}c}
\toprule
Adapter & Dataset  & $AR_{VSD}$ & $AR_{MSSD}$ & $AR_{MSPD}$ & $AR$\\ \hline
\multirow{+8}{*}{$M_A$} &LM  &96.0 & 95.7 & 95.7 &95.8  \\  
&LMO  &68.7 & 68.2 & 68.2 &68.4   \\  
 & TLESS    &85.9 & 85.1 & 85.8 &  85.6  \\ 
 & TUDL  &97.6 & 96.4 & 95.4 & 96.2  \\ 
 & ITODD  &69.2 & 68.5 & 68.4 & 68.7 \\ 
 & HB &    92.7 & 91.6 & 92.5 & 92.3\\ 
  & YCB &83.9 & 83.4 & 84   & 83.8\\ \cline{2-6}
  &average&84.9&84.1&84.3&\textbf{84.4}\\\hline
  \multirow{+8}{*}{$M^s_A$} &LM  & 93.4   & 92.7    & 92.6 & 92.9  \\  
  &LMO  &  59.7  &  59.3   &  59.2& 59.4 \\  
 & TLESS  & 78.8 &   78.7      & 78.8 &  78.8 \\ 
 & TUDL  &   95.6      &  95.3 & 95.5&95.5 \\ 
 & ITODD    &   56.7      &  56.7  & 56.5&56.6 \\ 
 & HB &    85.7     & 85.3   &85.6& 85.5\\ 
  & YCB &     76.5    & 76.4   & 76.6&76.5\\ 
  \cline{2-6}
  &average&78.1&77.8&77.8&77.9\\
  \bottomrule
\end{tabular}
\end{adjustbox}

\end{center}
\end{table}
\clearpage \newpage

\begin{table}[t]
\begin{center}
\caption{The impact of different RKHS kernels defined in Sec. 5.3 in the main paper on LM and five BOP core datasets.\label{tab:kernels supp}}
\begin{adjustbox}{max width=\columnwidth}
\begin{tabular}{ccccccV{1}r}
\toprule
\multirow{2}{*}{Kernels}&
\hspace{8 pt} Trainable \hspace{8 pt} & 
\multirow{2}{*}{Dataset}  & 
\multirow{2}{*}{$AR_{VSD}$} & 
\multirow{2}{*}{$AR_{MSSD}$} & 
\multirow{2}{*}{$AR_{MSPD}$\hspace{8 pt}}  & 
\multirow{2}{*}{\hspace{8 pt} $AR$}\\ 

& Weights ($w$) & & & & & \\
\hline
\multirow{+8}{*}{Linear}& \multirow{8}{*}{\xmark}&LM &85.4&84.2&84.4&84.7  \\  
&&LMO  & 56.9   &  56.3   & 55.4&56.2  \\  
& & TLESS &  74.3&	73.8&	74.2&	74.1 \\ 
& & TUDL  &  88.2&	87.3&	84.3&	86.6 \\ 
& & ITODD  &  45.2&	44.7&	45.3&	45.1 \\ 
& & HB &      79.3&	78.6&	79.2&	79.0\\ 
&  & YCB &    72.2&	70.5&	71.3&	71.3\\ \clineB{3-7}{1}
&  & average& 71.6&	70.8&	70.6&	71.0\\\clineB{1-7}{1}
\multirow{+8}{*}{RBF} & \multirow{8}{*}{\xmark} &LM  &  85.3&84.1&84.3&84.6\\  
&  &LMO  & 57.1   &  56.3   & 55.2&56.2     \\ 
& & TLESS &  73.7&	73.1&	73.6&	73.5   \\ 
& & TUDL  &  90.3&	89.7&	89.9&	90.0 \\ 
& & ITODD  &  52.7&	51.9&	52.5&	52.4 \\ 
& & HB &      80.2&	79.6&	79.3&	79.7 \\ 
&  & YCB &    72.6&	71.8&	71.5&	72.0 \\ \clineB{3-7}{1}
&  &average&  73.4&	72.9&	73.2&	73.2 \\\hlineB{3}
\multirow{+8}{*}{Linear}&\multirow{8}{*}{\checkmark} &LM  &96.0 & 95.7 & 95.7 &95.8  \\  
&&LMO  &68.7 & 68.2 & 68.2 &68.4   \\  
& & TLESS    &85.9 & 85.1 & 85.8 &  85.6  \\ 
& & TUDL  &97.6 & 96.4 & 95.4 & 96.2  \\ 
& & ITODD  &69.2 & 68.5 & 68.4 & 68.7 \\ 
& & HB &    92.7 & 91.6 & 92.5 & 92.3\\ 
&  & YCB &83.9 & 83.4 & 84   & 83.8\\ \clineB{3-7}{1}
&  & average&84.9&84.1&84.3&\textbf{84.4}\\\clineB{1-7}{1}
  \multirow{+8}{*}{RBF} &\multirow{8}{*}{\checkmark}  &LM  &  94.7  &  94.5   & 94.4 & 94.5 \\  
&  &LMO  & 57.1   &  56.3   & 55.2&56.2   \\  
& & TLESS    &     83.7    & 82.9 & 83.3&83.3   \\ 
& & TUDL  & 97.4   & 95.8 & 94.9&95.1 \\ 
& & ITODD    &  66.7       &  63.6  &64.2&  64.7\\ 
& & HB &    85.6     & 83.9   & 84.8&84.8\\ 
&  & YCB &    82.8     &  82.6  & 82.8&82.7\\ \clineB{3-7}{1}
&  &average&82.5&81.3&81.5&81.6\\
\bottomrule
\end{tabular}
\end{adjustbox}

\end{center}
\end{table}
\clearpage\newpage

\begin{table}[t]
\begin{center}
\caption{The impact of different metrics defined in Sec. 5.3 in the main paper for $M_A$ on LM and five BOP core datasets.\label{tab:metrics supp}}
\begin{adjustbox}{max width=\columnwidth}
\begin{tabular}{cccccV{1}r}
\toprule
Metrics & \hspace{4 pt} Dataset \hspace{4 pt}  & $AR_{VSD}$ & $AR_{MSSD}$ & $AR_{MSPD}$ \hspace{4 pt} & \hspace{8 pt} $AR$\\ \hline
\multirow{+8}{*}{MMD} &LM  &96.0 & 95.7 & 95.7 &95.8  \\  
&LMO  &68.7 & 68.2 & 68.2 &68.4   \\  
& TLESS    &85.9 & 85.1 & 85.8 &  85.6  \\ 
& TUDL  &97.6 & 96.4 & 95.4 & 96.2  \\ 
& ITODD  &69.2 & 68.5 & 68.4 & 68.7 \\ 
& HB &    92.7 & 91.6 & 92.5 & 92.3\\ 
& YCB &83.9 & 83.4 & 84   & 83.8\\ \clineB{2-6}{1}
 & average&84.9&84.1&84.3&\textbf{84.4}\\\hline
\multirow{+7.5}{*}{KL} 
  &LM &90.2&90.3&90.5& 90.3\\  
\multirow{+7.5}{*}{Div} 
 &LMO  &  63.4  &  62.2   & 63.6&63.1    \\  
 & TLESS    &   79.8      & 79.7 & 80.2&79.9   \\ 
 & TUDL  &     93.4    & 93.2  & 92.9&93.2 \\ 
 & ITODD    &   62.1      & 62.2   &62.1&62.1  \\ 
 & HB &    84.6     & 84.3   & 84.7&84.5\\ 
 & YCB &    72.6     &  72.5  & 72.3&72.4\\ \clineB{2-6}1
 & average&78.0&77.8&78.0&77.9\\\hline
\multirow{+7.5}{*}{Wass} 
  &LM  &  92.3&91.7&92.2&92.1  \\  
\multirow{+7.5}{*}{Distance}
  &LMO  &  64.7&64.2&64.7&64.5    \\  
& TLESS    &   82.3      & 82.2 &  81.9&82.1  \\ 
& TUDL  &    95.7     & 95.6  &95.5&95.6  \\ 
& ITODD    &    65.7     &  64.9  & 65.6&65.4 \\ 
& HB &    89.3     &  89.1  &89.5&89.3 \\ 
& YCB &    76.5     & 76.3   & 76.6&76.4\\\clineB{2-6}{1}
& average&80.9&80.6&80.9&80.8\\ 
\bottomrule
\end{tabular}
\end{adjustbox}

\end{center}
\end{table}
\clearpage\newpage
\begin{table}[t]
\begin{center}
\caption{The impact of different training strategies described in Sec. 5.2 in the main paper on LM and five BOP core datasets.\label{tab:train strategy supp}}
\begin{adjustbox}{max width=\columnwidth}
\begin{tabular}{cccccV{1}r}
\toprule
Training Type \hspace{8 pt} & Dataset  & $AR_{VSD}$ & $AR_{MSSD}$ & $AR_{MSPD}$\hspace{4 pt} & \hspace{8 pt}$AR$\\ \hline
\multirow{+8}{*}{Mixed} &LM  &96.0 & 95.7 & 95.7 &95.8  \\  
&LMO  &68.7 & 68.2 & 68.2 &68.4   \\  
 & TLESS    &85.9 & 85.1 & 85.8 &  85.6  \\ 
 & TUDL  &97.6 & 96.4 & 95.4 & 96.2  \\ 
 & ITODD  &69.2 & 68.5 & 68.4 & 68.7 \\ 
 & HB &    92.7 & 91.6 & 92.5 & 92.3\\ 
  & YCB &83.9 & 83.4 & 84   & 83.8\\ \clineB{2-6}{1}
   & average&84.9&84.1&84.3&\textbf{84.4}\\\hlineB{3}
  \multirow{+8}{*}{Sequential} 
  &LM  &  95.4&95.2&95.3&95.3  \\  
  &LMO  & 65.3   &  65.2   & 65.2&65.2    \\  
 & TLESS    &    85.7     & 85.3 &  85.6&85.5  \\ 
 & TUDL  &    97.6     &  96.4 & 95.4&96.2 \\ 
 & ITODD    &    63.4     &  62.2  & 62.3&62.6 \\ 
 & HB &    88.6     &  88.4  & 88.3&88.4\\ 
  & YCB &     79.8    &  79.5  & 80.0&79.8\\ \clineB{2-6}{1}
  &average&82.3&81.7&81.7&81.9\\
  \bottomrule
\end{tabular}
\end{adjustbox}

\end{center}
\end{table}
\clearpage\newpage

\section{Category Wise Performance}
\label{sec:Category Wise Performances S}
The category-wise ADD(S), ADD-S AUC, and ADD(S) AUC results
for LINEMOD-Occlusion, LINEMOD, and YCB-Video are shown in Tables~\ref{tab:OccLinemodFull}, 
~\ref{tab:LinemodFull},
and Table~\ref{tab:YCBVideoFull}.
RKHSPose outperforms on average against all self-supervised methods and most of the object categories.

\begin{table*}[h!]
\begin{center}
\caption{LMO accuracy results of self-supervised 6DOF PE methods:
accuracy of RKHSPose for non-symmetric objects is evaluated with ADD, and for
symmetric objects
(annotated with~$^{*}$) is evaluated with ADD-S. All the syn + real image methods use real images without GT labels. The methods annotated with~$^{**}$ are the TexPose~\cite{chen2023texpose} re-implementation.\label{tab:OccLinemodFull}
}
\begin{adjustbox}{max width=\textwidth}
\begin{tabular}{cccccccccc|c}
\cline{3-10}
\multicolumn{2}{c|}{}
&
\multicolumn{7}{c}{Object} & \multirow{2}{*}{hole-}  
& \multicolumn{1}{c}{}  
\\\cline{1-2}\cline{11-11}
Mode&Method  & ape  & can  & cat  & driller & duck & eggbox$^{*}$ & glue$^{*}$ & puncher  \hspace{6 pt} &\hspace{2 pt} Mean \\\hlineB{2}
syn&GDR~\cite{wang2021gdr}&44&83.9&49.1&88.5&15&33.9&75&34&52.9\\\hline
\multirow{+6}{*}{syn + real}&Self6D~\cite{wang2020self6d}&13.7&43.2&18.7&32.5&14.4&\textbf{57.8}&54.3&22&32.1\\
\multirow{+6}{*}{images}&Sock et al.~\cite{sock2020introducing}&12&27.5&12&20.5&23&25.1&27&35&22.8\\
&DSC~\cite{yang2021dsc}&9.1&21.1&26&33.5&12.2&39.4&37&20.4&24.8\\
&SMOC-Net~\cite{tan2023smoc}& 60.0&\underline{94.5}&\underline{59.1}&\textbf{93.0}&37.2&48.3&\textbf{89.3}&25.0&63.3\\
&Self6D++~$^{**}$~\cite{wang2021occlusion,chen2023texpose}&59.4&\textbf{96.5}&\textbf{60.8}&92&30.6&\underline{51.1}&88.6&38.5&64.7\\
&TexPose~\cite{chen2023texpose}&\underline{60.5}&93.4&56.1&92.5&\underline{55.5}&46&82.8&\underline{46.5}&66.7\\
&Ours&\textbf{62.7}&93.5&58.2&92.5&\textbf{58.7}&48.2&\underline{88.7}&\underline{46.5}&\underline{68.6}\\
&Ours+ICP&\textbf{62.7}&93.7&58.2&\underline{92.7}&\textbf{58.7}&48.3&\underline{88.7}&\textbf{46.7}&\textbf{68.7}\\\hline

\end{tabular}
\end{adjustbox}

\end{center}
\end{table*}
\begin{sidewaystable*}[t]
%
\begin{center}
\caption{ LINEMOD Accuracy Results of self-supervised 6DOF PE methods: Non-symmetric objects are evaluated with ADD, and
symmetric objects
(annotated with~$^{*}$) are evaluated with ADD-S
\label{tab:LinemodFull}. All the syn + real image methods use real images without GT labels. The methods annotated with~$^{**}$ are the TexPose~\cite{chen2023texpose} re-implementations. DeepIM annotated with~$^{\#}$ is the self6D++ re-implementation.
}

\begin{tabular}{ccccccccccccccc|cc}
\cline{3-15}
\multicolumn{2}{c|}{}
&
\multicolumn{13}{c|}{
Object}& \multicolumn{1}{c}{} \\
\multicolumn{2}{c|}{}
&
\multicolumn{1}{c}{}  &
\multicolumn{1}{c}{bench-}  &
\multicolumn{7}{c}{}  &
\multicolumn{1}{c}{hole-}  &
\multicolumn{3}{c|}{} & \multicolumn{1}{c}{} \\
\cline{1-2}\cline{16-16}
Mode      & Method     & ape & vise & camera  & can  & cat  & driller & duck & eggbox$^{*}$ & glue$^{*}$ & puncher & iron & lamp & phone & mean \\\hlineB{3}
\multirow{3}{*}{syn}  &AAE~\cite{sundermeyer2018implicit}&4.0&20.9&30.5&35.9&17.9&24.0&4.9&81.0&45.5&17.6&32.0&60.5&33.8&31.4\\
&MHP~\cite{Manhardt_2019_ICCV}&11.9&66.2&22.4&59.8&26.9&44.6&8.3&55.7&54.6&15.5&60.8&-&34.4&38.8\\
&DeepIM$^{\#}$~\cite{li2018deepim,wang2021occlusion} &85.8&93.1&99.1&\textbf{99.8}&98.7&\textbf{100.0}&61.9&93.5&93.3&32.1&\textbf{100.0}&\underline{99.1}&94.8&88.0\\\hline

\multirow{7}{*}{syn +} &DSC~\cite{yang2021dsc}&31.2&83.0&49.6&56.5&57.9&73.7&31.3&96.0&63.4&38.8&61.9&64.7&54.4&58.6\\
\multirow{7}{*}{real}&Self6D~\cite{wang2020self6d}&38.9&75.2&36.9&65.6&57.9&67.0&19.6&99.0&94.1&16.2&77.9&68.2&50.1&58.9\\
\multirow{7}{*}{image}&GDR$^{**}$~\cite{wang2021gdr,chen2023texpose}&85.0&\textbf{99.8}&96.5&99.3&93.0&\textbf{100.0}&65.3&\textbf{99.9}&98.1&73.4&86.9&\textbf{99.6}&86.3&91.0\\
&Sock et al.~\cite{sock2020introducing}&37.6&78.6&65.6&65.6&52.5&48.8&35.1&89.2&64.5&41.5&80.9&70.7&60.5&60.6\\
&Self6D++$^{**}$~\cite{wang2021occlusion,chen2023texpose}&75.4&94.9&97.0&99.5&86.6&98.9&68.3&99.0&96.1&41.9&99.4&98.9&94.3&88.5\\
&SMOC-Net~\cite{tan2023smoc}&85.6&96.7&97.2&99.9&95.0&\textbf{100.0}&76.0&98.3&\underline{99.2}&45.6&\underline{99.9}&98.9&94.0&91.3\\
&TexPose~\cite{chen2023texpose}&80.9&99&94.8&\underline{99.7}&92.6&97.4&83.4&94.9&93.4&79.3&99.8&98.3&78.9&91.7\\

&DPODv2~\cite{shugurov2021dpodv2}&80.0&\underline{99.7}&\textbf{99.2}&\underline{99.6}&95.1&98.9&79.5&99.6&99.8&72.3&99.4&96.3&\underline{96.8}&93.5\\
&Ours &\underline{90.2}&\underline{99.7}&\underline{99.1}&\textbf{99.8}&\underline{96.2}&99.2&\underline{86.3}&\underline{99.8}&\textbf{99.8}&\underline{80.3}&99.6&98.8&\textbf{97.2}&\underline{95.8}\\
&Ours+ICP &\textbf{90.3}&\underline{99.7}&\underline{99.1}&\textbf{99.8}&\textbf{96.4}&\underline{99.3}&\textbf{86.5}&\underline{99.8}&\textbf{99.8}&\textbf{80.7}&99.6&98.8&\textbf{97.2}&\textbf{95.9}\\\hline

syn + &SO-Pose~\cite{Di_2021_ICCV}&-&-&-&-&-&-&-&-&-&-&-&-&-&96.0\\
real&Ours &\underline{92.3}&\underline{99.7}&\textbf{99.3}&\textbf{99.8}&\textbf{97.5}&99.2&\underline{90.2}&\underline{99.8}&\textbf{99.8}&\underline{84.3}&99.6&98.8&\underline{97.2}&\underline{96.7}\\
labels&Ours+ICP &\textbf{92.7}&\underline{99.7}&\textbf{99.3}&\textbf{99.8}&\textbf{97.5}&99.2&\textbf{90.7}&\underline{99.8}&\textbf{99.8}&\textbf{84.5}&99.6&98.8&\textbf{97.3}&\textbf{96.8}\\
\bottomrule
\end{tabular}

\end{center}

\end{sidewaystable*}
\clearpage\newpage

\begin{sidewaystable*}[t]
\begin{center}
\caption{YCB ADD-S and ADD(S) AUC~\cite{posecnn} results of self-supervised 6DoF PE methods: Non-symmetric objects are evaluated with ADD AUC, and
symmetric objects
(annotated with~$^{*}$) are evaluated with ADD-S AUC. 
To the best of our knowledge, self6d++~\cite{wang2021occlusion} is the only self supervision method that provided YCB results.
\vspace{\baselineskip}
\label{tab:YCBVideoFull}
}
\vspace{0.6cm}
\begin{adjustbox}{max width=\textwidth}
\begin{tabular}{cccccccccccccccccccccccc|c}
\multicolumn{1}{c}{}&\multicolumn{1}{c}{}&&
\begin{rotate}{60}
\hspace{-0.45 cm} 002 master
\end{rotate}
&
\begin{rotate}{60}
\hspace{-0.45 cm} 003 cracker
\end{rotate}
&
\begin{rotate}{60}
\hspace{-0.45 cm} 004 sugar
\end{rotate}
&
\begin{rotate}{60}
\hspace{-0.45 cm} 005 tomato  
\end{rotate}
&
\begin{rotate}{60}
\hspace{-0.45 cm} 006 mustard  
\end{rotate}
&
\begin{rotate}{60}
\hspace{-0.45 cm} 007 tuna fish  
\end{rotate}
&
\begin{rotate}{60}
\hspace{-0.45 cm} 008 pudding 
\end{rotate}
&
\begin{rotate}{60}
\hspace{-0.45 cm} 009 gelatin
\end{rotate}
&
\begin{rotate}{60}
\hspace{-0.45 cm} 010 potted  
\end{rotate}
&
\begin{rotate}{60}
\hspace{-0.45 cm} 011 banana  
\end{rotate}
&
\begin{rotate}{60}
\hspace{-0.45 cm} 019 pitcher 
\end{rotate}
&
\begin{rotate}{60}
\hspace{-0.45 cm} 021 bleach 
\end{rotate}
&
\begin{rotate}{60}
\hspace{-0.45 cm} 024 bowl$^{*}$ 
\end{rotate}
&
\begin{rotate}{60}
\hspace{-0.45 cm} 025 mug 
\end{rotate}
&
\begin{rotate}{60}
\hspace{-0.45 cm} 035 power 
\end{rotate}
&
\begin{rotate}{60}
\hspace{-0.45 cm} 036 wood 
\end{rotate}
&
\begin{rotate}{ 60}
\hspace{-0.45 cm} 037 scissors 
\end{rotate}
&
\begin{rotate}{60}
\hspace{-0.45 cm} 040 large 
\end{rotate}
&
\begin{rotate}{60}
\hspace{-0.45 cm} 051 large 
\end{rotate}
&
\begin{rotate}{60}
\hspace{-0.45 cm} 052 extra large$^{*}$ 
\end{rotate}
&
\begin{rotate}{60}
\hspace{-0.45 cm} 061 foam 
\end{rotate}
& \multicolumn{1}{c}{}\\

\multicolumn{1}{c}{\hspace{8 pt}Mode\hspace{8 pt}} & \multicolumn{1}{c}{\hspace{8 pt}Metric\hspace{8 pt}} & \multicolumn{1}{c}{Method}                         & 
\begin{rotate}{60}
\hspace{0.8 cm} {chef can}
\end{rotate}

& 
\begin{rotate}{60}
\hspace{0.8 cm}
{box} 
\end{rotate}
&
\begin{rotate}{60}
\hspace{0.8 cm}{box} 
\end{rotate}
&
\begin{rotate}{60}
\hspace{0.8 cm}{soup can} 
\end{rotate}
&
\begin{rotate}{60}
\hspace{0.8 cm}{bottle} 
\end{rotate}
&
\begin{rotate}{60}
\hspace{0.8 cm}{can} 
\end{rotate}
&
\begin{rotate}{60}
\hspace{0.8 cm}{box} 
\end{rotate}
&
\begin{rotate}{60}
\hspace{0.8 cm}{box} 
\end{rotate}
&
\begin{rotate}{60}
\hspace{0.8 cm}{meat can} 
\end{rotate}
&
&
\begin{rotate}{60}
\hspace{0.8 cm}{base} 
\end{rotate}
&
\begin{rotate}{60}
\hspace{0.8 cm}{cleanser} 
\end{rotate}
&
&
&
\begin{rotate}{60}
\hspace{0.8 cm}{drill} 
\end{rotate}
&
\begin{rotate}{60}
\hspace{0.8 cm}{block$^{*}$} 
\end{rotate}
&
&
\begin{rotate}{60}
\hspace{0.8 cm}{marker} 
\end{rotate}
&
\begin{rotate}{60}
\hspace{0.8 cm}{clamp$^{*}$} 
\end{rotate}
&
\begin{rotate}{60}
\hspace{0.8 cm}{clamp}
\end{rotate}
&
\begin{rotate}{60}
\hspace{0.8 cm}{brick$^{*}$} 
\end{rotate}
&
{Mean}   
\\ \hline 
\multirow{4}{*}{syn+}&\multirow{2}{*}{ADD-S}&Self6D++~\cite{wang2021occlusion}&\underline{88.8}&94.2            &95.8            &90.8            &98.6            &97.5            &\underline{98.4}&94.0            &89.3            &\underline{98.5}&98.9            &93.5            &89.1            &94.1            &95.2            &78.3            &69.2            &87.5            &79.2            &87.3            &\underline{95.5}&91.1\\
\multirow{4}{*}{real}&\multirow{2}{*}{AUC}&Ours                               &88.7            &\underline{94.7}&\underline{96.2}&\underline{92.2}&\underline{99.5}&\underline{98.2}&98.3            &\underline{95.2}&\underline{92.7}&98.4            &\underline{99.1}&\underline{94.2}&\underline{92.3}&\underline{95.2}&\underline{95.5}&\underline{81.2}&\underline{71.3}&\underline{89.2}&\underline{83.4}&\underline{90.2}&\underline{95.5}&\underline{92.4}\\
\multirow{4}{*}{images}&&Ours+ICP                                             &\textbf{88.9}   &\textbf{94.9}   &\textbf{96.4}   &\textbf{92.3}   &\textbf{99.7}   &\textbf{98.5}   &\textbf{98.5}   &\textbf{95.5}   &\textbf{92.9}   &\textbf{98.6}   &\textbf{99.2}   &\textbf{94.4}   &\textbf{92.5}   &\textbf{95.4}   &\textbf{95.7}   &\textbf{81.4}   &\textbf{71.5}   &\textbf{89.4}   &\textbf{83.6}   &\textbf{90.4}   &\textbf{95.7}&\textbf{92.6}\\\cline{2-25}
&\multirow{2}{*}{ADD(S)}&Self6D++~\cite{wang2021occlusion}                    &8.4             &84.9            &88.0            &79.4            &\underline{92.7}&89.7            &93.9            &83.9            &\underline{75.7}&91.8            &92.1            &84.5            &89.1            &81.4            &84.2            &78.3            &45.2            &74.6            &79.2            &87.3            &\underline{95.5}&80.0\\
&\multirow{2}{*}{AUC}&Ours                                                    &\underline{13.7}&\underline{86.2}&\underline{91.3}&\underline{83.2}&\underline{92.7}&\underline{92.3}&\underline{94.3}&\underline{84.2}&\textbf{76.3}   &\underline{93.7}&\underline{94.3}&\underline{86.0}&\underline{92.3}&\underline{83.2}&\underline{86.3}&\underline{81.2}&\underline{62.3}&\underline{75.6}&\underline{83.4}&\underline{90.2}&\underline{95.5}&\underline{82.8}\\
&&Ours+ICP                                                                    &\textbf{13.8}   &\textbf{86.5}   &\textbf{91.5}   &\textbf{83.3}   &\textbf{93.6}   &\textbf{92.5}   &\textbf{94.5}   &\textbf{84.5}   &\textbf{76.3}   &\textbf{93.9}   &\textbf{94.5}   &\textbf{86.1}   &\textbf{92.5}   &\textbf{83.4}   &\textbf{86.5}   &\textbf{81.4}   &\textbf{62.5}   &\textbf{75.7}   &\textbf{83.6}   &\textbf{90.4}   &\textbf{95.7}&\textbf{83.0}\\\hline

\multirow{4}{*}{syn+}&\multirow{2}{*}{ADD-S}&Self6D++~\cite{wang2021occlusion}&93.8            &98.8            &\textbf{99.6}   &95.4            &\textbf{100.0}   &\underline{99.9}&63.3            &92.9            &91.1            &93.0            &99.3            &91.2            &87.2            &\textbf{96.4}   &\underline{99.7}&68.6            &78.9            &93.0            &81.7            &86.9            &94.3            &90.7\\
\multirow{4}{*}{real}&\multirow{2}{*}{AUC}&Ours                               &\underline{95.4}&98.8            &99.2            &\underline{96.3}&99.6             &99.8            &\underline{67.2}&\underline{93.5}&\underline{94.3}&\underline{95.2}&\underline{99.5}&\underline{93.7}&\underline{92.3}&\underline{96.3}&99.6            &\underline{71.2}&\underline{81.2}&\underline{94.3}&\underline{84.2}&\underline{90.2}&\underline{94.8}&\underline{92.2}\\
\multirow{4}{*}{labels}&&Ours+ICP                                             &\textbf{95.7}   &99.0            &\underline{99.5}&\textbf{96.5}   &\underline{99.8} &\textbf{100.0}  &\textbf{67.5}   &\textbf{93.7}   &\textbf{94.5}   &\textbf{95.5}   &\textbf{99.7}   &\textbf{93.9}   &\textbf{92.5}   &\textbf{96.4}   &\textbf{99.8}   &\textbf{71.4}   &\textbf{81.4}   &\textbf{94.4}   &\textbf{84.3}   &\textbf{90.4}   &\textbf{95.2}   &\textbf{92.4}\\\cline{2-25}
&\multirow{2}{*}{ADD(S)}&Self6D++~\cite{wang2021occlusion}                    &56.7            &92.8            &\underline{95.0}&90.5            &94.7             &\underline{97.0}&42.1            &84.7            &78.2            &80.5            &98.7            &81.9            &87.2            &86.6            &93.6            &68.6            &61.3            &81.7            &81.7            &86.9            &94.3            &82.6\\
&\multirow{2}{*}{AUC}&Ours                                                    &\underline{62.3}&\underline{95.3}&94.9            &\underline{93.2}&\underline{95.2} &\underline{97.0}&\underline{50.2}&\underline{87.2}&\underline{81.2}&\underline{83.2}&\underline{99.1}&\underline{83.2}&\underline{92.3}&\underline{87.2}&\underline{95.2}&\underline{71.2}&\underline{74.3}&\underline{82.7}&\underline{84.2}&\underline{90.2}&\underline{94.8}&\underline{85.4}\\
&&Ours+ICP                                                                    &\textbf{62.7}   &\textbf{95.6}   &\textbf{95.2}   &\textbf{93.4}   &\textbf{95.5}    &\textbf{97.2}   &\textbf{50.5}   &\textbf{87.5}   &\textbf{81.4}    &\textbf{83.3}   &\textbf{99.2}   &\textbf{83.4}   &\textbf{92.5}   &\textbf{87.3}   &\textbf{95.4}   &\textbf{71.4}   &\textbf{74.4}   &\textbf{82.8}   &\textbf{84.3}   &\textbf{90.4}   &\textbf{95.2}   &\textbf{85.6}\\
\hline

\end{tabular}
\end{adjustbox}

\end{center}

\end{sidewaystable*}
\clearpage\newpage
%
%
\clearpage
\bibliographystyle{splncs04}
\bibliography{main}
\end{document}